\pdfoutput=1
\documentclass[10pt,logo,copyright]{nvidiatechreport}
\usepackage[authoryear,sort&compress,round]{natbib}

\usepackage[utf8]{inputenc} 
\usepackage[T1]{fontenc}    

\usepackage{parskip}        
\usepackage{url}            
\usepackage{hyperref}       
\usepackage{booktabs}       
\usepackage{amsfonts}       
\usepackage{nicefrac}       
\usepackage{microtype}      
\usepackage{xcolor}         
\usepackage[dvipsnames]{xcolor} 
\usepackage{graphicx}
\usepackage{animate}        
\usepackage{subcaption}
\usepackage{tabularx}
\usepackage{makecell}
\usepackage{adjustbox}
\usepackage{setspace}
\newcolumntype{M}[1]{>{\centering\arraybackslash}m{#1}}
\usepackage{float}
\usepackage{tikz}
\usetikzlibrary{positioning,shapes,arrows}
\usepackage{amsmath,amsfonts,bm, bbm,leftindex}
\usepackage{multirow}
\usepackage{comment}
\usepackage{gensymb}
\usepackage{lipsum}
\usetikzlibrary{arrows.meta, positioning, fit}
\usepackage[para]{threeparttable}
\usepackage{tikz}
\usetikzlibrary{tikzmark}

\setcitestyle{numbers,square,comma}

\definecolor{darkred}{rgb}{0.7, 0.0, 0.0}

\usepackage{pifont}
\usepackage{wrapfig}

\usepackage[nameinlink]{cleveref}
\crefname{equation}{Eq.}{Eqs.}
\crefname{figure}{Fig.}{Figs.}
\crefname{section}{Sec.}{Sec.}
\crefname{appendix}{App.}{App.}
\crefname{table}{Tab.}{Tabs.}
\crefname{algorithm}{Algo}{Algo}
\crefname{thm}{Thm}{Thm}
\Crefname{thm}{Thm}{Thm}
\crefname{prop}{Prop}{Prop}
\usepackage{longtable}

\usepackage{amsfonts}
\usepackage{bm}
\usepackage{amsmath}
\usepackage{mathtools}
\usepackage{multirow}
\usepackage{booktabs}
\usepackage{caption}
\usepackage{enumitem}
\usepackage{wrapfig,lipsum,booktabs}
\usepackage{caption}
\usepackage{bbm}
\usepackage{multirow}
\usepackage{pifont}
\usepackage[linesnumbered,ruled,vlined]{algorithm2e}

\SetCommentSty{mycommfont}
\SetAlCapFnt{\small} 
\usepackage{etoc}
\SetAlgorithmName{Algo}{Algo}{}

\renewcommand{\paragraph}[1]{{\vspace{1mm}\noindent \bf #1}.}

\newcommand{\link}{https://nvlabs.github.io/GEAR-SONIC/}

\newcommand{\rewardfunc}{\mathcal{R}}

\newcommand{\goalstate}{{\bs{s}^{\text{g}}_t}}

\newcommand{\selfstate}{{\bs{s}^{\text{p}}_t}}

\newcommand{\state}{{\bs{s}_t}}
\newcommand{\action}{{\bs{a}_t}}

\newcommand{\mpjpe}{E_\text{mpjpe}}

\newcommand{\acc}{E_\text{acc}}
\newcommand{\vel}{E_\text{vel}}
\newcommand{\success}{\text{Succ}}

\newcommand{\simp}{{\bs{{q}}_{t}}}

\newcommand{\simv}{{\bs{\dot{q}}_{t}}}

\newcommand{\simav}{{\bs{{\omega}}_{t}}}

\newcommand{\bs}[1]{\boldsymbol{#1}}

\newcommand{\shortname}{\texttt{SONIC}\xspace}

\newcommand{\secref}[1]{the \nameref{#1} section}

\newcommand{\crefnames}[3]{%
  \@for\next:=#1\do{%
    \expandafter\crefname\expandafter{\next}{#2}{#3}%
  }%
}

\title{SONIC: Supersizing Motion Tracking for Natural Humanoid Whole-Body Control}

\author{
	 Zhengyi Luo$^{\dagger}$, Ye Yuan$^{\dagger}$, Tingwu Wang$^{\dagger}$, Chenran Li$^{\dagger}$, Fernando Casta\~neda$^{\dagger}$, Sirui Chen$^{\ast}$,
     Zi-Ang Cao$^{\ast}$, Jiefeng Li$^{\ast}$, David Minor$^{\ast}$, Qingwei Ben$^\ast$, Jinhyung Park$^\ast$, David Sami$^\ast$, Zi Wang$^{\ast}$, Xingye Da$^\ast$,
     Runyu Ding, Cyrus Hogg, Lina Song, Edy Lim, Eugene Jeong, Tairan He, Haoru Xue, Wenli Xiao,
     Simon Yuen, Jan Kautz, Yan Chang, Umar Iqbal, Linxi "Jim" Fan$^{\ddagger}$, Yuke Zhu$^{\ddagger}$ \\
	\small NVIDIA, 2788 San Tomas Expressway, Santa Clara, CA, USA \\
    \small$^\dagger$ Co-first Authors, \small$^\ast$ Core Contributors, \small$^\ddagger$ Equal Advising \\
    \small \tt{\href{\link}{\link}}
}

\begin{document}
\maketitle
\newcommand{\tingwu}[1]{\textcolor{blue}{ [TW: \textbf{#1}]}}
\newcommand{\zen}[1]{\textcolor{green}{ [Zen: \textbf{#1}]}}
\newcommand{\ye}[1]{\textcolor{teal}{ [Ye: \textbf{#1}]}}

\begin{abstract}

    Despite the rise of billion-parameter foundation models trained across thousands of graphical processing units (GPUs), similar scaling gains have not been shown for humanoid control. Current neural controllers for humanoids remain modest in size, target a limited set of behaviors, and are trained on a handful of GPUs. We show that scaling model capacity, data, and compute yields a generalist humanoid controller capable of natural, robust whole-body movements. We position motion tracking as a scalable task for humanoid control, leveraging dense supervision from diverse motion-capture data to acquire human motion priors without manual reward engineering. We build a foundation model for motion tracking by scaling along three axes: network size (1.2M to 42M parameters), dataset volume (100M+ frames from 700 hours of motion capture), and compute (21k GPU hours). Beyond demonstrating the benefits of scale, we further show downstream utility through a real-time kinematic planner that bridges motion tracking to tasks such as navigation, enabling natural and interactive control, as well as a unified token space that supports virtual reality (VR) teleoperation and vision-language-action (VLA) models with a single policy. Through this interface, we demonstrate autonomous VLA-driven whole-body loco-manipulation requiring coordinated hand and foot placement. Scaling motion tracking exhibits favorable properties: performance improves steadily with compute and data diversity, and learned policies generalize to unseen motions, establishing motion tracking at scale as a practical foundation for humanoid control.

\end{abstract}

\abscontent

\begin{figure}[th]
    \centering
    \includegraphics[width=\textwidth]{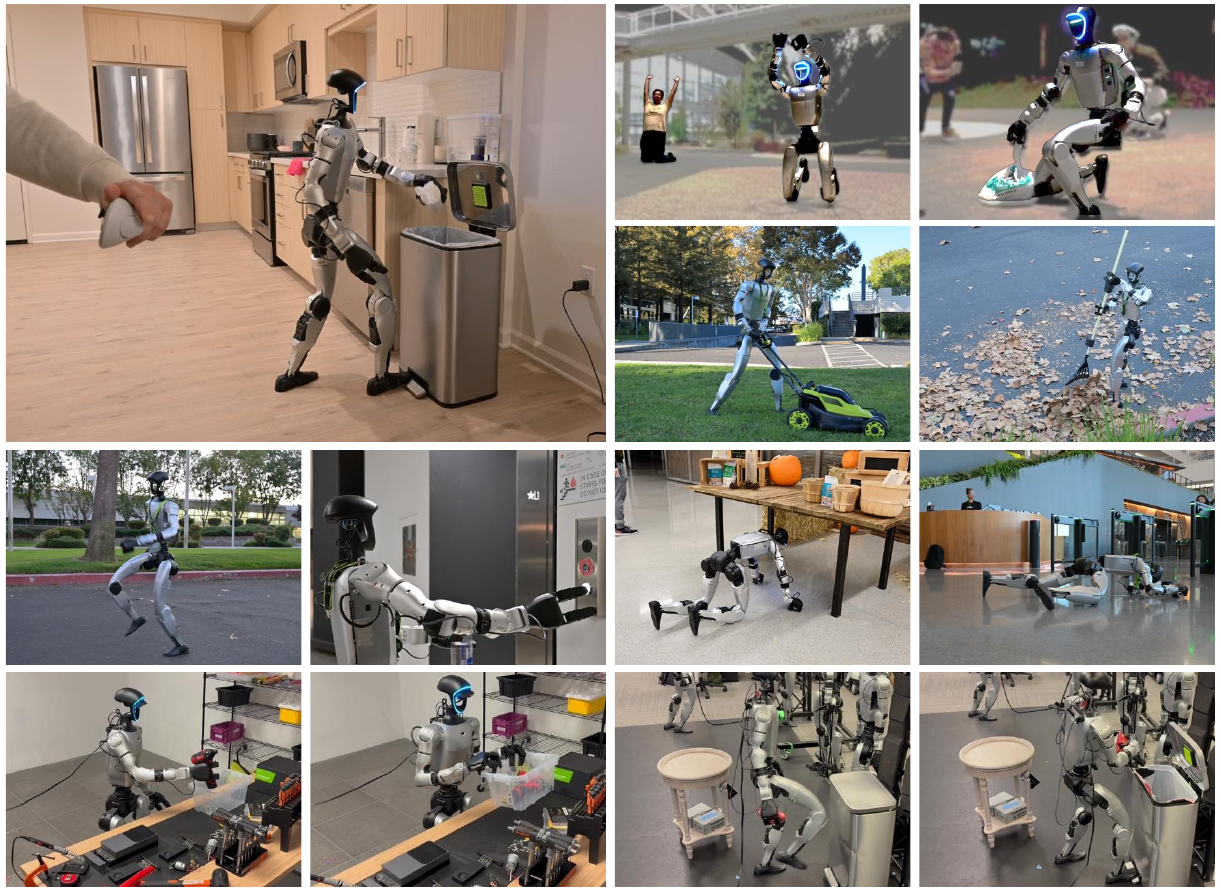}
    \caption{\shortname enables diverse humanoid tasks through a universal control policy that handles diverse input modalities and control interfaces.}
    \label{fig:main_result}
\end{figure}

\section{Introduction}





Over the past decade, artificial intelligence has scaled rapidly: the generative pretrained transformer (GPT) \citep{achiam2023gpt} family of models is trained on 25,000+ graphical processing units (GPUs) with trillions of tokens; video and image-generation models \citep{ramesh2022hierarchical,rombach2022high,brooks2024video,ho2022imagen, blattmann2023svd} leverage thousands of GPUs processing billions of images. These foundation models have shown a consistent pattern: scale unlocks emergent capabilities, generalization, and robustness that smaller models cannot achieve \citep{kaplan2020scaling,hoffmann2022chinchilla,wei2022emergent}. Yet for sim-to-real humanoid control, similar scaling gains have not been achieved. State-of-the-art humanoid control policies are often small neural networks, such as three-layer multilayer perceptrons (MLPs), trained on a few GPUs for a single task. Manually engineered reward terms are designed per task and do not generalize across behaviors, fundamentally limiting the scalability of these approaches.

Why hasn't humanoid control scaled? The fundamental issue is task selection. Tasks like locomotion require extensive reward engineering for each scenario---walking forward naturally provides little signal for dancing \citep{he2025asap}, getting up from the ground \citep{he2025learning,huang2025learning}, or teleoperation \citep{ben2025homie,li2025amo,ze2025twist}. Each new capability demands redesigned rewards and objectives, making scaling up difficult. Generative imitation methods such as AMP \citep{peng2021amp}, ASE \citep{peng2022ase}, and CALM \citep{tessler2023calm} provide a unified objective by combining matching motion distributions and simple task-specific rewards, but prior work has shown that their discriminator-based training signal is prone to mode collapse as the motion dataset grows in size and diversity \citep{tessler2024maskedmimic, luo2023universal}. Even if we identify a scalable objective that can learn diverse behaviors, a second challenge emerges: how do we support the diverse range of real-world applications? A desired humanoid controller should handle teleoperation, goal-directed tasks, navigation, and even vision-language commands \citep{ahn2022saycan,brohan2022rt1,brohan2023rt2,team2023openx,ma2024droid}. Building a system that scales while remaining flexible for different task specifications is non-trivial.

In this work, we address both challenges by identifying motion tracking as the scalable foundational task for humanoid control. Motion tracking leverages human motion capture data, which provides dense, frame-by-frame supervision without reward engineering. Critically, humanoids benefit from decades of motion capture research---datasets covering walking, running, dancing, sports, and object interactions already exist at scale \citep{Mahmood_2019_ICCV,punnakkal2021babel,Li_2021_ICCV}. Although there exist prior works on motion tracking \citep{liao2025beyondmimic, zhang2025track, zeng2025behavior, yin2025unitrackerlearninguniversalwholebody, chen2025gmt, Luo2023PerpetualHC, wang2020unicon, he2025hover, he2024omnih2o}, they are mostly limited to showing whole-body motion tracking results on \textbf{training data} and have not demonstrated many downstream tasks beyond motion tracking or navigation. We supersize physics-based motion tracking to 100 million frames (at 50\,Hz) with 128-GPU training, achieving universal tracking capabilities across diverse human behaviors while maintaining real-time performance. In addition, we show how such a motion tracker can be applied to meaningful downstream tasks, and introduce two key contributions. First, we develop a universal kinematic motion generation system for interactive control, enabling goal-directed tasks such as interactive locomotion and game-like character control through kinematic planning in motion space. Second, we design a universal token space that unifies heterogeneous motion sources, including VR teleoperation, vision-language-action models, and generative motion models that convert video, text, and music into motion \citep{tevet2022mdm,zhang2023t2mgpt,kocabas2020vibe,li2021bailando}, within a single control interface.

We propose \textit{\textcolor{BrickRed}{S}upersizing m\textcolor{BrickRed}{O}tion tracking for \textcolor{BrickRed}{N}atural humano\textcolor{BrickRed}{I}d \textcolor{BrickRed}{C}ontrol} (\shortname), a framework that enables natural humanoid control across a wide range of applications (\href{https://youtu.be/2BWl2uAO5lc}{Movie~S1}). We achieve high-precision teleoperation and interactive control capabilities, including running, jumping, and crawling with natural human-like movement. Leveraging our universal token space, our controller can directly map estimated whole-body human motion to humanoid control signals, bypassing the need for explicit retargeting at runtime. We integrate our tracking policy with multi-modal human motion generation models, supporting video, text, and music control. Furthermore, we show that teleoperation data collected through our system can be used to train vision-language-action foundation models \citep{bjorck2025gr00t}, establishing a complete pipeline from motion tracking at scale to foundation model-based humanoid control, including tasks that demand simultaneous hand-foot coordination for whole-body loco-manipulation. These results validate that large-scale motion tracking serves as a practical foundational task for diverse real-world humanoid applications.

Our contributions are threefold. First, we identify motion tracking as a scalable foundational task for humanoid control, demonstrating that it exhibits favorable scaling properties with both compute and data diversity; we scale up humanoid control to 21k GPU hours and 100 million frames of motion sequences, achieving universal tracking capabilities across diverse human behaviors. Second, we introduce a real-time kinematic motion generator for interactive control and a universal token space with specialized encoders for robot, human, and hybrid motion inputs, all mapped into a shared quantized representation. Third, we provide a comprehensive evaluation demonstrating humanoid control scaling trends, zero-shot transfer to unseen motions, robust sim-to-real deployment on physical humanoid robots, and successful integration with foundation models; we show that the universal token space enables VLA-driven whole-body loco-manipulation, including tasks requiring coordinated hand grasping and precise foot placement, across five real-world tasks.

\section{Results}

\begin{figure}[htbp]
    \centering
    \includegraphics[width=\textwidth]{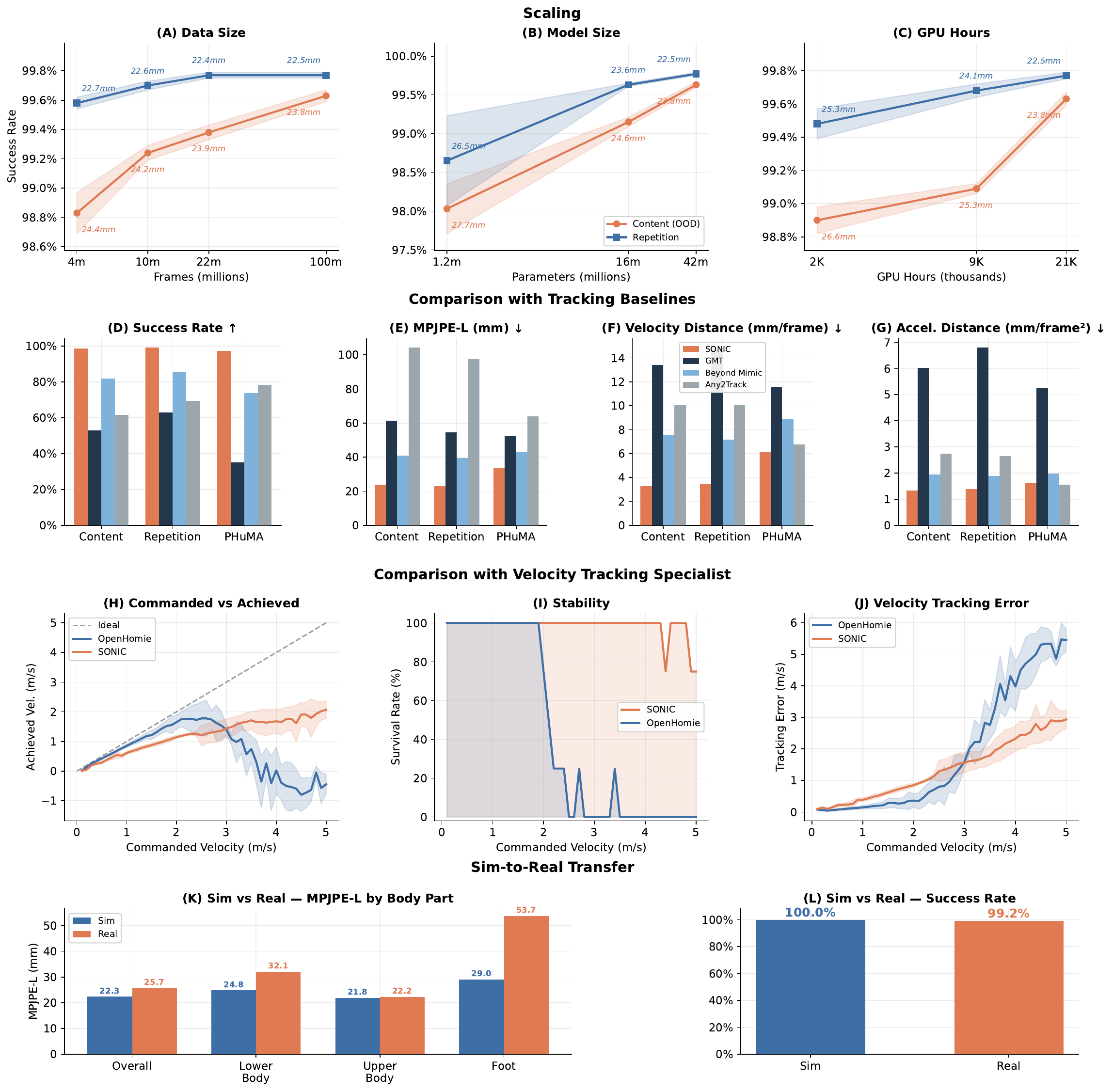}
    \caption{\footnotesize\textbf{Scaling and benchmarking of \shortname for universal motion tracking.} \textbf{(A to C)} Effect of scaling data size, model size, and compute on test-content (unseen motion content, OOD) and test-repetition (held-out takes of seen motion types). Lines and shaded bands: mean $\pm$1 s.d.\ over $n{=}6$ evaluations per configuration. Success rate improves along all three axes (Welch's $t$-test, all $P\leq0.011$). \textbf{(D to G)} Comparison with baselines on test-content, test-repetition, and PHUMA~\citep{lee2025phuma} ($N{=}7{,}016$, $9{,}395$, and $68{,}326$ motions): (D)~success rate (\shortname vs.\ each baseline, Welch's $t$-test, all $P<0.001$); (E)~MPJPE-L, (F)~velocity distance, and (G)~acceleration distance, reported as means over successfully tracked motions. \textbf{(H to J)} Velocity tracking vs.\ the OpenHomie specialist~\citep{ben2025homie}: (H)~commanded vs.\ achieved velocity, (I)~survival rate ($197/200$ vs.\ $86/200$ runs, Welch's $t$-test, $P<0.001$), and (J)~tracking error (paired $t$-test across commanded velocities, $P<0.001$ in both velocity regimes); mean $\pm$1 s.d.\ over $n{=}4$ directions. \textbf{(K and L)} Sim-to-real transfer: (K)~MPJPE-L by body part and (L)~success rate on the same $N{=}124$ motions (sim $124/124$ vs.\ real $123/124$).}
    \label{fig:compare_baselines}
\end{figure}

We used the Unitree G1 humanoid \citep{unitreeHumanoidRobot} to demonstrate our supersizing humanoid motion tracking framework, \shortname. We demonstrated \shortname's motion tracking capabilities, interactive kinematic motion planning, multi-modal control, teleoperation, and loco-manipulation, shown in \Cref{fig:main_result}.

\subsection{Motion Tracking}
\label{sec:res:tracking}

\shortname, trained on 100 million frames of motion over 21k GPU hours (128 GPUs over 7 days), exhibited strong generalization to \textbf{unseen} motions. In this section, we evaluated the generalization capabilities of our tracker on large-scale, unseen motion datasets in simulation and the real world.

\paragraph{Metrics} We employed a comprehensive set of pose-based and physics-based metrics to measure motion imitation performance. The primary measure was the success rate ($\success$), where a motion imitation was deemed unsuccessful if the humanoid deviated too far from the reference motion trajectory. We further reported the local (root-relative) mean per-joint position error (MPJPE-L) $\mpjpe$ (in millimeters, mm), computed over 14 body links (pelvis, knees, ankles, torso, elbows, wrists)~\citep{liao2025beyondmimic}, quantifying the accuracy of the imitation in the robot's local frame. To assess physical fidelity, we also calculated differences in acceleration ($\acc$, mm/frame$^2$) and velocity ($\vel$, mm/frame) between the simulated humanoid and the reference human motion.

\paragraph{Evaluation Protocol and Dataset}
We evaluated on three test sets (\Cref{tab:dataset_splits}). From our motion-capture dataset, we constructed two held-out splits. The first, \textbf{test-content} (7,016 clips, 15 hours), evaluated generalization to \emph{unseen motion content} and contained 182 sub-categories entirely absent from training. The second, \textbf{test-repetition} (9,395 clips, 12 hours), evaluated robustness to \emph{new performances and repetitions} of known motion types. For baseline comparisons, we additionally evaluated on \textbf{PHUMA}~\citep{lee2025phuma}, a publicly available dataset of 68,326 motions from a different retargeting pipeline. Motion imitation was considered unsuccessful if the humanoid's root height or end-effector height deviated by more than 0.25\,m from the reference. Unlike prior work that defines success via global root position error \citep{Luo2023PerpetualHC} (for example, a 0.5\,m threshold), our tracker performed \emph{local} motion tracking rather than following a global trajectory. Our metric, similar to \citep{liao2025beyondmimic}, captured the physically meaningful failure modes such as falling. See \secref{sec:method:data} for dataset details.

\paragraph{Scaling Up Motion Tracking}
In \Cref{fig:compare_baselines}(A--C), we analyzed scaling along three axes. For data size, we compared training on 4M, 10M, 22M, and 100M frames (corresponding to 20k, 50k, 110k, and 317k motion clips); smaller subsets were created by uniformly sampling across motion sub-categories to preserve distribution diversity. For model size, we scaled from 1.2M to 16M to 42M parameters. For compute, we trained on 2, 4, and 16 nodes (16, 32, and 128 GPUs), all to 50k iterations, yielding approximately 2k, 9k, and 21k GPU hours. All evaluations used Isaac Lab~\citep{mittal2025isaaclab}, with models trained 50k iterations (when performance usually plateaus). Scaling yielded consistent improvements on both test-content (out-of-distribution, OOD) and test-repetition: the largest model achieved 99.6\% success with 23.8\,mm MPJPE-L on test-content, compared to 98.0\% success with 27.7\,mm for the smallest (1.2M). Gains were most pronounced on OOD motions, confirming that scale improves generalization. For compute, more GPUs yielded better asymptotic performance at the same iteration count, as larger batch sizes improved optimization stability. Shaded regions denote $\pm$1 standard deviation across 6 evaluation checkpoints. Visualizations of out-of-distribution test motions, including successful and failed tracking cases, are provided in the Supplementary Materials (\Cref{fig:success_failure}). We also demonstrated robustness to strong external perturbations in \Cref{fig:appendix_push}.

\paragraph{Comparison with Other Motion Trackers}
We compared against state-of-the-art trackers: GMT~\citep{chen2025gmt}, Any2Track~\citep{zhang2025track}, and BeyondMimic~\citep{liao2025beyondmimic}. These baselines were trained on different source datasets (Any2Track and BeyondMimic on LaFAN; GMT on AMASS); for single-motion trackers like BeyondMimic, we retrained for multi-motion tracking using the publicly released code. To maximize protocol consistency, all methods were evaluated in MuJoCo~\citep{todorov2012mujoco} under the same termination criterion defined before (\Cref{fig:compare_baselines}(D--G)). This comparison primarily reflects cross-dataset generalization and scaling effects rather than a fully data-matched benchmark, as the baselines were trained on different source data and retargeting pipelines than \shortname. Under this setting, \shortname achieved 98.5\%/99.2\%/97.2\% success on test-content/test-repetition/PHUMA~\citep{lee2025phuma}, compared to 82.0\%/85.4\%/73.8\% for BeyondMimic and 61.6\%/69.4\%/78.5\% for Any2Track. On tracking accuracy, \shortname achieved 23.7\,mm MPJPE-L, a 42\% reduction over BeyondMimic (40.9\,mm). The 97.2\% success rate on PHUMA is particularly notable because PHUMA aggregates motions from video-based pose estimation with a different retargeting pipeline~\citep{lee2025phuma}, making it substantially more out-of-distribution than our own held-out splits.

\paragraph{Comparison with Specialist Baseline}
To show that a universal tracker can match or exceed specialist controllers, we compared \shortname against OpenHomie~\citep{ben2025homie}, a state-of-the-art single-task locomotion controller optimized for upper body inverse kinematics control and lower-body velocity tracking. We evaluated both systems on sim-to-sim velocity tracking in MuJoCo across the 0--5\,m/s command range (\Cref{fig:compare_baselines}(H--J)). \shortname achieved a 98.5\% overall survival rate compared to OpenHomie's 43.0\%. OpenHomie's survival rate collapsed beyond $\sim$2\,m/s, dropping below 20\% beyond $\sim$2.5\,m/s, whereas \shortname maintained near-100\% stability up to $\sim$4\,m/s. Notably, OpenHomie was specifically designed and trained for locomotion; \shortname's universal policy, trained on diverse whole-body motion data combined with a motion generator, outperformed it.

\paragraph{Real-World Evaluation}
We assessed the real-world performance of \shortname by deploying it on 124 diverse motion sequences (\Cref{fig:real_eval_motions}). As presented in \Cref{fig:compare_baselines}(K--L), our policy achieved motion imitation in the real world that closely matched its simulation results. The real-world policy achieved 99.2\% success rate compared to 100\% in simulation, with an overall MPJPE-L of 25.7\,mm (vs.\ 22.3\,mm in sim). The sim-to-real gap was smallest for the upper body (22.2\,mm real vs.\ 21.8\,mm sim) and largest for the feet (53.7\,mm vs.\ 29.0\,mm), reflecting the difficulty of precise foot placement under real-world contact dynamics.

\begin{figure}[htbp]
    \centering
    \includegraphics[width=0.98\textwidth]{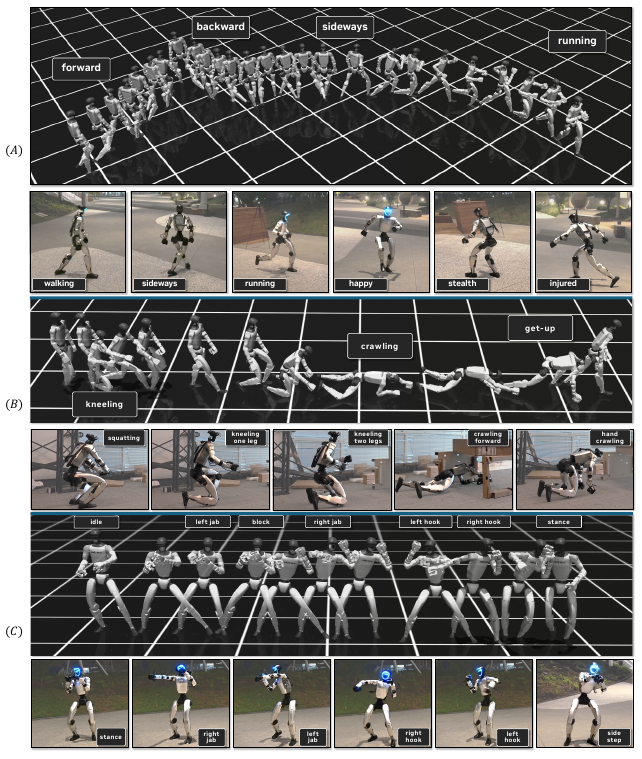}
    \caption{\textbf{Interactive whole-body control with \shortname.} A single universal control policy, driven by the real-time kinematic planner, executes diverse user-commanded behaviors on the Unitree G1 in both simulation and the real world. \textbf{(A)}~Navigation across velocities, directions, and locomotion styles; \textbf{(B)}~squatting, kneeling, crawling, and getting up at arbitrary pelvis heights; \textbf{(C)}~continuous boxing. Each row is a temporally ordered frame sequence.}
    \label{fig:result_kinematic}
\end{figure}

\subsection{Interactive Motion Control}

\label{sec:res:planner}

In this section, we demonstrated the scalability and robustness of \shortname in whole-body, real-time interactive control tasks (\href{https://nvlabs.github.io/GEAR-SONIC/\#interactive-kinematic-planner}{Movie~S2}).
We presented a kinematic runtime generative motion planner that guided the robot's policy through user interaction.
Our approach employed an autoregressive framework that continually regenerated future kinematic motions conditioned on the previous states and incoming user commands.
For each planning step, the model generated motion segments lasting between 0.8\,s and 2.4\,s,
where the duration was automatically determined by the neural planner to maximize flexibility and robustness.
The planner achieved inference times under 5\,ms on a standard laptop and ${\sim}$12\,ms on a Jetson Orin GPU.
Replanning was triggered as frequently as every 100\,ms, or immediately when user commands were updated, ensuring highly responsive control.

\shortname supported a variety of applications, including navigation control with arbitrary velocity, direction, and style commands; interactive entertainment tasks such as boxing; and locomotion skills such as squatting, crawling, and kneeling, which are useful for downstream applications like teleoperation, because both our kinematic planner and tracking policy were trained on the same large-scale dataset.
Utilizing the scalable nature of \shortname, we noted that all the applications above were specified after training, without retraining the planner or the tracking policy.

For navigation control, \shortname accepted velocity commands ranging from 0.0\,m/s to 6.0\,m/s,
as well as arbitrary direction commands spanning 0 to 360 degrees.
We note that 6.0\,m/s represents the upper bound of the \emph{command} range; the actual achievable velocity is limited by the tracker's capabilities and is filtered by a critically damped spring model (\secref{sec:method:planner}). The actual commanded-vs-achieved velocity analysis is presented in \Cref{fig:compare_baselines}(H).

\shortname achieved scalable, responsive, and robust navigation control and supported different styles such as drunken walking, injured walking, happy walking, and stealth walking, as shown in \Cref{fig:result_kinematic}(A).
The capacity of the model to generate robust in-betweening motions---the task of smoothly transitioning between motion keyframes or segments---showed the flexibility of \shortname,
and the potential for more natural human-robot interaction.

\shortname further extended its versatility to interactive entertainment tasks such as boxing,
as illustrated in \Cref{fig:result_kinematic}(C).
Existing academic solutions~\citep{starke2021neural, won2021control} and industrial approaches~\citep{unitree_boxing}
often utilize a limited collection of boxing clips, requiring a switch between multiple expert models or action labels.
This approach leads to discontinuous, unnatural transitions and even pauses.
\shortname enabled much more fluid, responsive, and natural motion generation,
retaining the robot's full freedom of movement throughout the task.

To enable downstream manipulation or navigation in confined environments, skills such as squatting, kneeling, and crawling are essential.
As illustrated in \Cref{fig:result_kinematic},
\shortname supported squatting, kneeling, and crawling. For squatting and kneeling, we
allowed the pelvis height to be smoothly controlled from 0.3\,m to 0.8\,m.
For navigation in especially tight spaces, \shortname also enabled crawling.
The robot could move omnidirectionally using its elbows and knees at velocities from 0.0\,m/s to 0.5\,m/s.

\begin{figure}[htbp]
    \centering

    \includegraphics[width=0.95\textwidth]{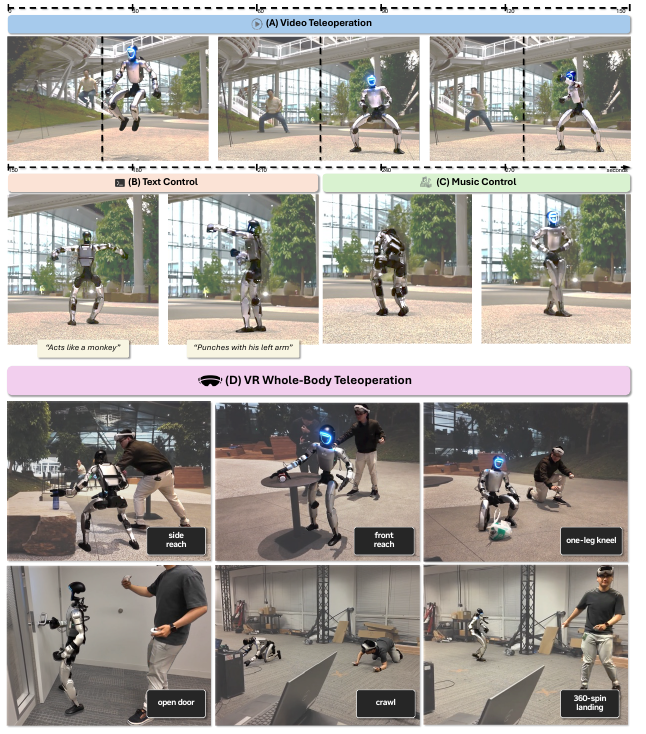}
    \caption{\textbf{Multi-modal, real-time control with \shortname.} A single universal control policy is driven by heterogeneous input modalities. \textbf{(A)}~Video teleoperation: full-body motion is estimated from a monocular RGB video stream and tracked in real time. \textbf{(B)}~Text control: target motions are generated from natural-language prompts. \textbf{(C)}~Music control: target motions are generated from music, conditioned on melodic and rhythmic structure. \textbf{(D)}~VR whole-body teleoperation: an operator drives the robot through a range of dynamic whole-body skills. Each panel shows a temporally ordered sequence of frames.}
    \label{fig:result_genmo}
\end{figure}

\subsection{Video Teleoperation and Multi-Modal Control}
\label{sec:res:multi-modal}
As shown in \Cref{fig:result_genmo}, \shortname also supported a real-time, multi-modal control framework, enabled by our universal control policy (\href{https://nvlabs.github.io/GEAR-SONIC/\#multi-modal-control}{Movie~S3}). A unified motion generation system based on the Generalist Model for human motion (GEM)~\citep{genmo2025}---a single model that performs both human motion estimation and generation---was designed to generate human motions from three input modalities: videos, natural-language commands, and music audio. The multi-modal motion generation system coordinated smooth transitions among modalities.

\paragraph{Video Teleoperation}
For video control, the system supported both pre-recorded clips and live monocular webcam streams. Human motion was estimated at $\geq$60 frames per second (fps), enabling interactive teleoperation without specialized motion-capture hardware. Video control provided a higher-fidelity specification of pose and timing, yielding a precise imitation of movements.

\paragraph{Music and Text Control}
For text control, the system accepted natural-language prompts and synthesized target motions at $\geq$60\,fps. An interactive graphical interface supported free-form prompting at any time with immediate on-robot responses (for example, ``walk forward'', ``kick left foot'', or ``dance like a monkey''). For music control, GEM generated dance motions (which our tracking policy imitated) conditioned on melodic and rhythmic structure, tempo, and musical characteristics. Our system supported transitions between modalities. For example, users could initiate fine-grained control via video, switch to text for general control, and finally hand off to music for performance.

\subsection{VR-Based Teleoperation}
\label{sec:res:teleop}

We built two VR teleoperation interfaces on top of \shortname (\href{https://nvlabs.github.io/GEAR-SONIC/\#teleoperation}{Movie~S4}). \emph{Whole-body teleoperation} used a PICO headset, ankle trackers, and handheld controllers~\citep{zhao2025xrobotoolkit} to stream full-body poses in the Skinned Multi-Person Linear (SMPL)~\citep{SMPL:2015} format, encoded via the human motion encoder~$\bs{\mathcal{E}}_h$. \emph{3-point teleoperation} used only the headset and controllers (no ankle trackers), outputting three upper-body SE(3) poses (head, both wrists), hand joint angles, waist height, and a navigation command; the kinematic planner generated the lower body. Both interfaces used the same universal token space; video-based teleoperation is also supported (\secref{sec:res:multi-modal}). We used the VR teleoperation interfaces to collect teleoperation data for training VLA foundation models. Full interface details are provided in the Supplementary Materials (\secref{app:teleop_vla}).

\subsection{Foundation-Model-Driven Loco-manipulation}
\label{sec:vla_control}

\begin{figure}[htbp]
    \centering
    \includegraphics[width=\textwidth]{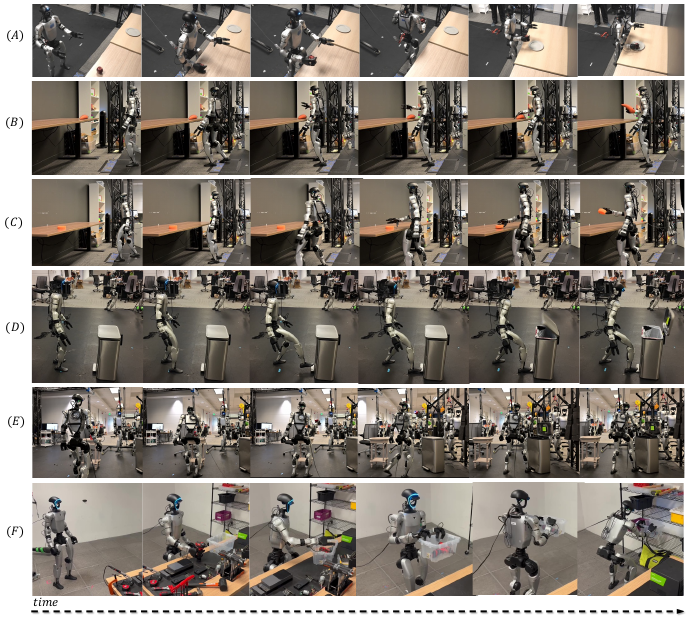}
    \caption{\footnotesize\textbf{VLA-driven loco-manipulation tasks.} Each row shows a temporally ordered sequence of frames (time proceeds left to right). \textbf{(A)} Apple-to-plate pick-and-place via the 3-point interface. \textbf{(B)} Carrot pickup. \textbf{(C)} Scrub pickup. \textbf{(D)} Open trash can by stepping on the pedal. \textbf{(E)} Soda can to trash can: pick up, navigate, step on pedal, and throw. \textbf{(F)} Drill and box relocation: pick up drill, place in box, and carry to shelf. See \Cref{tab:vla_tasks} for success rates.}
    \label{fig:VLA}
\end{figure}

We connected a GR00T N1.5 VLA model~\citep{gr00t-n1_5-blog, bjorck2025gr00t} to the universal token interface, enabling autonomous whole-body control (\Cref{fig:VLA}, \Cref{tab:vla_tasks}; \href{https://nvlabs.github.io/GEAR-SONIC/\#vla-foundation-model}{Movie~S5}). We evaluated the resulting VLA-driven system on five loco-manipulation tasks of increasing complexity. The first task (apple to plate) used the 3-point interface; the remaining four used the whole-body interface, where the VLA predicted a 78-dimensional action comprising a 64-dimensional universal motion token and 14-dimensional hand joint angles. We observed that predicting universal tokens produced smoother and safer behavior than predicting explicit SMPL poses, which resulted in jerky motions and poor directional control (see \secref{sec:ablations} for an ablation). All success rates are strict binary outcomes over 10--20 trials per task (\Cref{tab:vla_tasks}).

\textbf{Apple to plate} (3-point interface): The robot walked to a table, picked up an apple, and placed it on a plate. Trained on 300 trajectories, the policy achieved 90\% success over 20 trials.

\textbf{Object pickup} (carrot, scrub; whole-body interface): The robot walked to a table, located a target object, and grasped it under randomized table heights [24--30 inches (60.96--76.2\,cm)] and starting positions. Trained on 3,900 trajectories (300 per object, 13 objects), the policy achieved 75\% (carrot) and 95\% (scrub) success.

\textbf{Open trash can} (whole-body interface): The robot navigated to a trash can and stepped on a pedal to open the lid, requiring precise foot placement and dynamic single-leg balance under closed-loop VLA control. Trained on 200 trajectories, the policy achieved 70\% success. This task required the VLA to coordinate full-body dynamics, using feet as manipulators.

\textbf{Soda can to trash can} (whole-body interface): The most complex task, combining five sequential skills: walk to the table, pick up the can with one hand, navigate to the trash can, open the lid by stepping on the pedal with one foot while balancing on the other, and throw the can inside. This required simultaneous hand manipulation, foot manipulation, and dynamic balance within a single action sequence. Trained on 1,000 multi-object trajectories, the policy achieved 60\% success.

\textbf{Drill and box relocation} (whole-body interface): A multi-stage task where the robot picked up a drill, placed it in a box, and carried the box to a shelf with both hands. Trained on 300 trajectories, the policy achieved 70\% success.

Across all five tasks (10--20 trials each), the VLA achieved 75\% average success using the universal token action space. The soda-can and trash-can tasks illustrate autonomous whole-body loco-manipulation with coordinated hand and foot placement, a capability that would be difficult to realize with action spaces that decouple upper-body control from locomotion.

\begin{table}[htbp]
    \centering
    \footnotesize
    \setlength{\tabcolsep}{6pt}
    \begin{tabular}{llccc}
        \toprule
        \multicolumn{5}{l}{\textit{(A) VLA task success rates (universal motion token action space)}} \\
        \midrule
        \textbf{Task}                                   & \textbf{Interface}              & \textbf{Training Data}                       & \textbf{Trials}         & \textbf{Success}                   \\
        \midrule
        Apple to plate             & 3-point    & 300 trajs (single-obj)  & 20 & 90\%          \\
        Object pickup (carrot)     & whole-body & 3,900 trajs (multi-obj) & 20 & 75\%          \\
        Object pickup (scrub)      & whole-body & 3,900 trajs (multi-obj) & 20 & 95\%          \\
        Open trash can (foot)      & whole-body & 200 trajs               & 10 & 70\%          \\
        Soda can to trash can      & whole-body & 1,000 trajs (multi-obj) & 10 & 60\%          \\
        Drill and box relocation   & whole-body & 300 trajs               & 10 & 70\%          \\
        \midrule
        \textbf{Average (5 tasks)} &                                 &                                              &                         & \textbf{75\%} \\
        \bottomrule
    \end{tabular}

    \vspace{0.9em}

    \begin{tabular}{lccc}
        \toprule
        \multicolumn{4}{l}{\textit{(B) Action space: universal motion tokens vs.\ explicit SMPL poses}} \\
        \midrule
        \textbf{Task} & \textbf{FSQ Token} & \textbf{SMPL Poses} & \textbf{$\Delta$} \\
        \midrule
        Carrot pickup          & \textbf{75\%} & 60\% & +15          \\
        Open trash can (foot)  & \textbf{70\%} & 20\% & +50          \\
        Soda can to trash can  & \textbf{60\%} & 0\%  & +60          \\
        \midrule
        \textbf{Average}       & \textbf{68\%} & 27\% & \textbf{+42} \\
        \bottomrule
    \end{tabular}
    \caption{\textbf{Vision-language-action (VLA) control through the universal token interface.} \textbf{(A)} Task success rates. A GR00T N1.5 model, fine-tuned on teleoperated data, is evaluated across five whole-body loco-manipulation tasks (the object-pickup variants share one policy and are averaged as a single task for the 5-task mean); success is a strict binary outcome (no partial credit), measured over $n{=}10$--$20$ trials per task. \textbf{(B)} Action-space ablation: task completion success rate using universal motion tokens vs.\ explicit SMPL poses. The compact, structured FSQ token space is substantially easier for the VLA to learn, and the gap widens on more complex tasks (for example, 60\% vs.\ 0\% on soda-can-to-trash-can).}
    \label{tab:vla_tasks}
    \label{tab:vla_token_vs_smpl}
\end{table}

\clearpage

\subsection{Discussion}

We cast motion tracking as a scalable task for learning a single, versatile humanoid controller. By training \shortname on 100 million+ motion frames with up to 128 GPUs, we obtain a single policy that produces natural, robust whole-body behaviors across diverse conditions. Equally important, we build the practical system that makes tracking usable in real deployments: a real-time kinematic motion planner that converts intent into short-horizon reference motions, and a universal token space that unifies heterogeneous interfaces (teleoperation, video, text, and music) within one policy.

We observe consistent improvements as data, model capacity, and compute increase, with generalization to unseen motions in simulation and real-world deployments. These findings support motion tracking as a practical route to acquire broad, transferable whole-body priors without per-task reward engineering.

\paragraph{Why does motion tracking scale} We attribute the favorable scaling properties of motion tracking to its dense, per-frame supervisory signal. Each training frame provides an explicit target pose, so the learning signal remains informative as the dataset grows in size and diversity. This stands in contrast to adversarial imitation methods (AMP~\citep{peng2021amp}, ASE~\citep{peng2022ase}), where a discriminator must distinguish real from generated motions across the full distribution; as diversity increases, the discriminator's task becomes harder and its feedback less informative, leading to mode collapse~\citep{tessler2024maskedmimic, luo2023universal}. It also contrasts with task-specific reward engineering (for example, locomotion controllers such as OpenHomie~\citep{ben2025homie}), where each behavior requires a tailored objective that does not generalize. Our comparison with OpenHomie demonstrates this concretely: even on velocity tracking, \shortname's universal tracker achieves 98.5\% survival versus OpenHomie's 43.0\%, showing that data diversity benefits a universal tracker more than specialization benefits a narrow one. Furthermore, OpenHomie's velocity tracking performance plateaus when scaling beyond 8 GPUs (\Cref{fig:homie_scaling}), whereas \shortname continues to improve with additional compute.

Since the token space represents the full body, VLAs can control the entire kinematic chain, including the feet. Our VLA experiments demonstrate tasks requiring coordinated hand grasping and precise foot placement within a single action sequence.

Limitations include the lack of formal treatment of safety and energy efficiency for extended deployments. The tracker is robust to noisy planner output through domain randomization on motion commands during training and the critically damped spring model that filters unrealistic commands at deployment, but under more extreme conditions or very dynamic motions, the tracker may lose balance.

In summary, scaling motion tracking yields reliable, general whole-body control; pairing it with a planner, a universal token space, and an efficient onboard stack makes it usable as a system. We expect \shortname to serve as a practical foundation upon which higher-level perception and reasoning can be built to advance general-purpose humanoid autonomy.

\section{Materials and Methods}

\paragraph{Study Design}
We evaluated whether physics-based motion tracking scales favorably with data, model size, and compute for humanoid whole-body control. We trained policies in simulation (Isaac Lab~\citep{mittal2025isaaclab}) and tested them in both simulation and the real world. We systematically varied data size, model size, and compute (\secref{sec:res:tracking}) and evaluated on three held-out test sets with predefined splits (\Cref{tab:dataset_splits}): two from our dataset (test-content, test-repetition) and one external benchmark (PHUMA). Scaling curves report the mean $\pm$1 standard deviation over $n{=}6$ independent evaluations per configuration. Real-world evaluation covered 124 motion sequences (one trial per sequence; 123 completed successfully). Statistical procedures are detailed in \secref{sec:stats}. VLA task success rates (\Cref{tab:vla_tasks}) were measured over 10--20 trials per task. All training runs used the same hyperparameters (\Cref{tab:ppo_params}) and reward function (\Cref{tab:rl_rewards}).

\subsection{Humanoid Motion Dataset}
\label{sec:method:data}

Our motion dataset was built from a large-scale motion-capture collection. To promote diversity among participants, the collection includes a balanced mix of male and female performers. The dataset spans a broad spectrum of everyday human behaviors, including locomotion, daily activities, gesturing, and a diverse set of combat motions with varied stylistic expressions. Clip durations range from 1 to 180 seconds. In total, the collection covers thousands of unique motion behaviors, with most actions performed by multiple participants across multiple takes, providing rich intra- and inter-participant variation, as can be seen in \Cref{fig:dataset_panorama}. The source dataset contains approximately 700 hours of human motion. After retargeting to the Unitree G1 using General Motion Retargeting (GMR)~\citep{joao2025gmr} and PyRoki~\citep{kim2025pyroki}, we filtered out physically implausible motions (such as stair climbing and seated activities) that cannot be executed on the target robot, yielding 611 hours of training data (100+ million frames at 50\,Hz).

\paragraph{Dataset Diversity and Splits}
The dataset spans 33 motion categories (\Cref{tab:dataset_splits}), including basic and advanced locomotion, dance (hip-hop, Latin, vogue, fila), gestures, combat (sword, martial arts, magic), object manipulation (one-handed and two-handed at varying heights and object sizes), tool use (valves, levers, chainsaws, brooms), injured-gait, stylistic variations (drunk, zombie, stealth), role-play, and more. Each motion is captured from multiple participants and is mirrored, yielding paired left/right variants.
We constructed explicit train/test splits to enable rigorous evaluation (\Cref{tab:dataset_splits}). The \textbf{training set} covers 8,447 unique motion sub-categories (611 hours). The \textbf{test-content} split isolates \emph{unseen motion content} (sub-categories entirely absent from training), whereas \textbf{test-repetition} isolates \emph{unseen repetitions} of known content (different takes and actor performances).

\paragraph{Public Data Release}
A substantial portion of our motion-capture dataset has been publicly released as the BONES-SEED dataset~\citep{bones_seed}, available on Hugging Face. BONES-SEED contains 142,220 annotated motion sequences (288 hours) from 522 actors in SOMA~\citep{saito2025soma} and Unitree G1 formats, with natural language descriptions, temporal segmentation labels, and actor information.

\begin{table}[h!]
    \centering
    \footnotesize
    \setlength{\tabcolsep}{6pt}
    \begin{tabular}{lrrr}
        \toprule
                                                     & \textbf{Train} & \textbf{Test-Content} & \textbf{Test-Rep.} \\
        \midrule
        \multicolumn{4}{l}{\textit{Split statistics}}                                                              \\
        \quad Clips                                  & 317,189        & 7,016                 & 9,395              \\
        \quad Duration (hours)                       & 611            & 15                    & 12                 \\
        \quad Unique sub-categories                  & 8,447          & 182                   & 1,088              \\
        \quad Sub-cat.\ overlap w/ train             & ---            & 0\%                   & 100\%              \\
        \quad Clip overlap w/ train                  & ---            & 0\%                   & 0\%                \\
        \midrule
        \multicolumn{4}{l}{\textit{Main categories (33 total)}}                                                    \\
        \quad Locomotion (basic + adv.)              & 53,255         & 2,481                 & 2,683              \\
        \quad Gestures                               & 37,939         & 1,488                 & 1,125              \\
        \quad Acting / Roleplay                      & 68,742         & ---                   & 20                 \\
        \quad Combat (sword, martial arts, and more) & 50,162         & ---                   & ---                \\
        \quad Props / Object manip.                  & 14,513         & 701                   & 253                \\
        \quad Dance                                  & 9,689          & 504                   & 485                \\
        \quad Injured                                & 9,386          & 1,167                 & 528                \\
        \quad Action / Tool use                      & 9,920          & 228                   & 322                \\
        \quad Others (10+ main cat.)                 & 63,583         & 429                   & 890                \\
        \bottomrule
    \end{tabular}
    \caption{Dataset split statistics and main/sub-category distribution. Each \emph{main category} (such as Locomotion and Dance) contains many \emph{sub-categories} describing specific motion types (such as ``hip-hop slide'' and ``injured-leg jog''). \textbf{Test-content} evaluates generalization to \emph{unseen sub-categories}: its sub-categories are entirely absent from training. \textbf{Test-repetition} evaluates robustness to \emph{unseen performances}: all sub-categories overlap with training, but the specific clips are disjoint.}
    \label{tab:dataset_splits}
\end{table}

\subsection{Universal Humanoid Motion Tracking}
\label{sec:method:tracking}

\begin{figure}[htbp]
    \centering
    \includegraphics[width=\textwidth]{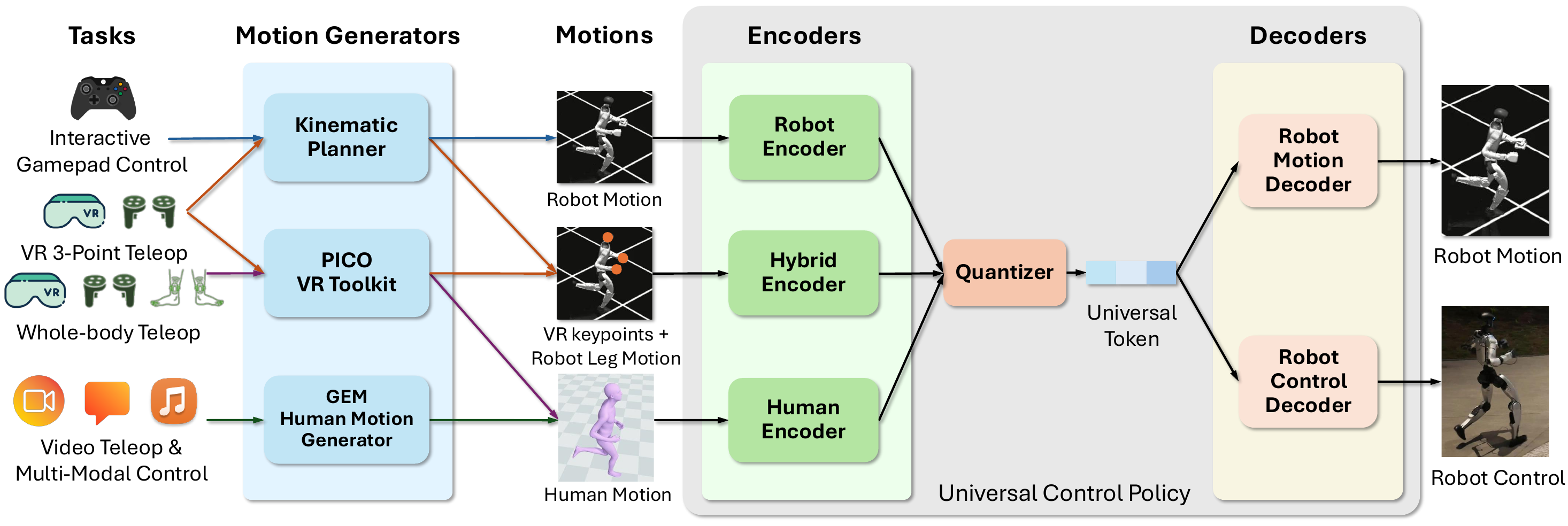}
    \caption{\textbf{\shortname enables universal humanoid motion tracking through a universal control policy that handles diverse motion commands and modalities.} Specialized encoders process robot, human, and hybrid motion commands into a universal token that drives robot control and motion decoders. This multi-encoder design supports diverse applications including gamepad control, VR teleoperation, whole-body teleoperation, and video teleoperation.}
    \label{fig:overview}
\end{figure}

\Cref{fig:overview} provides an overview of our approach, \shortname, a universal humanoid motion tracking framework that employs a unified control policy to track diverse motion commands from multiple input formats. A key innovation is its ability to seamlessly handle robot motion, human motion, and hybrid motion (combining upper-body keypoints with lower-body robot motions) through a shared latent representation. We used various motion generators (kinematic motion planner, VR motion generator, human motion generator (GEM)) to generate motion commands, which enable diverse applications including interactive gamepad control, VR 3-point teleoperation, whole-body teleoperation, video-based teleoperation, and multi-modal control from text and music.

\paragraph{Motion Tracking Formulation}
We formulated humanoid motion tracking as a Markov Decision Process ${\mathcal M}=\langle \mathcal{\bs S}, \mathcal{ \bs A}, \mathcal{ \bs T}, \rewardfunc, \gamma\rangle$, comprising state space, action space, transition function, reward function, and discount factor $\gamma$. We trained the policy using proximal policy optimization (PPO)~\citep{schulman2017proximal} to maximize the expected cumulative discounted return $\mathbb{E}\left[\sum_{t=1}^{T} \gamma^{t-1} r_{t}\right]$. Our environment design followed the general motion tracking formulation \citep{Luo2023PerpetualHC, chen2025gmt, zhang2025track, he2024omnih2o, liao2025beyondmimic}, and we adapted the well-tuned environmental settings from \citep{liao2025beyondmimic} as the basis for scaling up humanoid motion tracking.

\textit{States.} The state representation $\state$ comprises two components: proprioceptive sensing $\selfstate$ and motion command $\goalstate$. Proprioceptive information $\selfstate$ includes joint pose $\simp$, joint velocity $\simv$, root angular velocity $\simav$, gravity vector $\bs{g}_t$ in the root frame, and previous action $\bs{a}_{t-1}$. We concatenated a 10-step history of all proprioceptive quantities and actions into $\selfstate$, that is, $\selfstate \triangleq (\simp, \simv, \simav, \bs{g}_t, \bs{a}_{t-1})_{t-9:t}$, providing the policy with temporal context for anticipatory behavior. The motion command $\goalstate$ has three types: robot motion $\bs{g}_r$, human motion $\bs{g}_h$, or hybrid motion $\bs{g}_m$ (combining upper-body keypoints with lower-body robot motions), where we drop the subscript $t$ for brevity. All state quantities are expressed in the robot's local frame to ensure rotation invariance. We used the 6D rotation representation \citep{zhou2019continuity} throughout.

\textit{Actions.} The policy $\pi$ outputs target joint positions $\action$ as actions, which are tracked by proportional-derivative (PD) controllers at each joint. For PD gain settings, we followed prior art \cite{RaibertFarshidian2025WBC, liao2025beyondmimic} that has proven effective in training high-quality tracking policies.

\textit{Rewards.} We defined the reward as $r_t = \rewardfunc(\selfstate, \goalstate) + \mathcal{P}(\selfstate, \bs{a}_t)$, combining a tracking term that minimizes root and body-link pose and velocity errors (including an end-effector term on the head, wrists, and ankles) with penalty terms that encourage smooth, stable motion. The complete reward design is provided in Table~\ref{tab:rl_rewards}.

\textit{Domain Randomization.} To enhance robustness and generalization, we applied systematic domain randomization during training, randomizing physical parameters, periodically perturbing the robot's root velocities to simulate external pushes, and perturbing the target motion commands $\goalstate$. All randomization ranges are detailed in Table~\ref{tab:rand_runtime_eq}.

\paragraph{Universal Control Policy}
A main innovation of our tracking framework is its ability to accommodate multiple motion command types from different embodiments through a unified encoder-decoder architecture. Specialized encoders map heterogeneous human and robot motion inputs into a shared latent representation, which is quantized into a \textit{universal token} that drives a common robot control decoder; an auxiliary robot motion decoder facilitates feature learning and serves as an implicit human-to-robot retargeting module. This lets the policy learn from diverse motion sources, including human motion, despite morphological differences.

\textit{Encoders.} Three specialized encoders process distinct motion command types: the robot motion encoder $\bs{\mathcal{E}}_r$ encodes robot joint positions and velocities over $F_r$ future frames with a frame interval $\Delta t_r$, the human motion encoder $\bs{\mathcal{E}}_h$ encodes 3D human joint positions \citep{SMPL:2015} over $F_h$ future frames with a frame interval $\Delta t_h$, and the hybrid motion encoder $\bs{\mathcal{E}}_m$ encodes sparse upper-body keypoints (head and hands) of the current frame (for real-time upper-body tracking), combined with lower-body robot motion over $F_m$ future frames with a frame interval $\Delta t_m$. Multi-frame inputs enable anticipatory behavior and improve the robustness of the policy. All encoders are implemented as multi-layer perceptrons (MLPs; architecture details in Table~\ref{tab:policy_params}) that map commands $\bs{g}_r, \bs{g}_h, \bs{g}_m$ into a shared latent space, enabling aligned representations across input modalities.

\textit{Quantizer.} The encoded latent representation is quantized into a \textit{universal token} $\bs{z}$ using Finite Scalar Quantization (FSQ)~\citep{mentzer2023finite}, with two tokens, each a $D_z$-dimensional vector with $L_z$ quantization levels per dimension. We chose FSQ over the vector-quantized variational autoencoder (VQ-VAE)~\citep{van2017neural} for training stability under joint PPO optimization (\secref{app:tables}), and validated this and other design choices in \secref{sec:ablations}.

\textit{Decoders.} The \textit{universal token} $\bs{z}$ is decoded through two separate decoders. First, a robot control decoder $\bs{\mathcal{D}}_c$ transforms the \textit{universal token} into motor commands that control the robot's joints. $\bs{\mathcal{D}}_c$ takes as input the concatenation of the universal token $\bs{z}$ and the proprioceptive state $\selfstate$, that is, $\bs{a}_t = \bs{\mathcal{D}}_c(\bs{z}, \selfstate)$, where all quantities are expressed in the local frame as defined above. The same input representation is used identically during training in simulation and real-world deployment. Second, a robot motion decoder $\bs{\mathcal{D}}_r$ reconstructs the robot motion command, providing auxiliary supervision to improve the latent space and enhance feature learning.   $\bs{\mathcal{D}}_r$ takes only the universal token as input, that is, $\hat{\bs{g}}_r = \bs{\mathcal{D}}_r(\bs{z})$. Both decoders are implemented as MLPs (Table~\ref{tab:policy_params}).

\paragraph{Training}
We prepared synchronized motion data across all three command types. Each command type $\bs{g}_r, \bs{g}_h, \bs{g}_m$ is encoded via its respective encoder and quantized to produce universal tokens $\bs{z}_r, \bs{z}_h, \bs{z}_m$. For each token, the control decoder $\bs{\mathcal{D}}_c$ generates motor commands, whereas the motion decoder $\bs{\mathcal{D}}_r$ reconstructs the robot motion command. The total loss comprises:
\begin{align}
    \mathcal{L}                & = \mathcal{L}_{\text{ppo}} + \mathcal{L}_{\text{recon}} + \mathcal{L}_{\text{token}} + \mathcal{L}_{\text{cycle}}                                                                      \\
    \mathcal{L}_{\text{recon}} & = \left\| \bs{\mathcal{D}}_r(\bs{z}_r) - \bs{g}_r \right\|^2 + \left\| \bs{\mathcal{D}}_r(\bs{z}_h) - \bs{g}_r \right\|^2 + \left\| \bs{\mathcal{D}}_r(\bs{z}_m) - \bs{g}_r \right\|^2 \\
    \mathcal{L}_{\text{token}} & = \left\|\bs{z}_r - \bs{z}_h \right\|^2 + \left\|\bs{z}_r - \bs{z}_m \right\|^2 + \left\|\bs{z}_m - \bs{z}_h \right\|^2                                                                \\
    \mathcal{L}_{\text{cycle}} & = \left\| \bs{\mathcal{E}}_r(\bs{\mathcal{D}}_r(\bs{z}_h)) - \bs{z}_r \right\|^2
\end{align}
where $\mathcal{L}_{\text{ppo}}$ denotes the standard PPO loss. $\mathcal{L}_{\text{recon}}$ represents the reconstruction loss for the robot motion command across different input modalities. Notably, when the input command is human motion $\bs{g}_h$, the encoder-decoder acts as a retargeting pipeline from human to robot motion, and $\mathcal{L}_{\text{recon}}$ serves as a retargeting loss that enables learning from human motion data. $\mathcal{L}_{\text{token}}$ enforces pairwise alignment between all three encoder outputs, ensuring that the same motion produces similar tokens for robot motion, human SMPL poses, or hybrid teleop commands when the source motion is the same. $\mathcal{L}_{\text{cycle}}$ is a cycle consistency loss between the original robot token $\bs{z}_r$ and the token produced by re-encoding the reconstructed robot motion from the human token, that is, $\bs{\mathcal{E}}_r(\bs{\mathcal{D}}_r(\bs{z}_h))$. This loss further reinforces latent space coherence, ensuring that the translation from human to robot motion and back preserves the essential motion characteristics.

All four losses are optimized jointly in a single end-to-end training loop. We used asymmetric actor-critic training~\citep{pinto2018asymmetric}: the critic observes privileged simulation state (base linear velocity, full body link positions and orientations, and noise-free observations) during training, whereas the actor operates solely on deployment-available observations (noisy proprioceptive sensing and motion commands). The PPO loss updates the encoders, quantizer, and control decoder $\bs{\mathcal{D}}_c$ (as well as the critic network); the reconstruction, token alignment, and cycle consistency losses update the encoders and motion decoder $\bs{\mathcal{D}}_r$. Gradients propagate through the FSQ quantizer via straight-through estimation~\citep{mentzer2023finite}, allowing PPO to shape the encoder representations. In practice, the auxiliary losses regularize the latent space and stabilize rather than destabilize PPO optimization, and we observed no training instabilities from coupling quantization with RL across model scales.

We employed bin-based adaptive motion sampling that partitions the dataset into fixed-duration bins and weights sampling by capped failure rates, balancing targeted practice on challenging motions with uniform coverage. We trained using distributed training powered by \cite{accelerate} and \cite{vonwerra2022trl} across multiple compute nodes in Isaac Lab~\citep{mittal2025isaaclab}. Training hyperparameters are provided in Table~\ref{tab:ppo_params}.

\paragraph{Tasks and Applications}
The multi-encoder design enables diverse applications through the same policy: interactive gamepad control via the kinematic planner and $\bs{\mathcal{E}}_r$; VR whole-body and 3-point teleoperation via $\bs{\mathcal{E}}_h$ and $\bs{\mathcal{E}}_m$ respectively; VLA-driven autonomous control by predicting universal tokens (\secref{sec:vla_control}); and multi-modal control (video, text, music) via GEM~\citep{genmo2025} and $\bs{\mathcal{E}}_h$ (\secref{sec:method:genmo}).

\subsection{Generative Kinematic Motion Planner}
\label{sec:method:planner}


    Our generative kinematic motion planner is a large-scale latent generative model,
    trained on the same natural whole-body motion data as the motion tracking policy.
    At a high level, the planning process was formulated as an autoregressive motion in-betweening generation task, that is, generating the intermediate motion that smoothly connects the current pose to a target keyframe.
    The context keyframes capture historical robot states, such as joint positions and root positions,
    whereas target keyframes are either navigation guidance keyframes generated from user commands such as velocity, direction, and style,
    or skill-specific targets for actions such as squatting, crawling, and boxing.

\paragraph{Motion Representation}
During training, we sampled motion segments of length between 0.8\,s and 2.4\,s,
extracting the keyframes at both endpoints to serve as the context and target keyframes.
Our motion representation is mathematically equivalent to the humanoid pose configuration $q_t$ as introduced in \secref{sec:method:tracking}.
Specifically, we represented kinematic motion using the pelvis-relative joint positions and global joint rotations.
During training, we randomly rotated the training samples to enable planning in all initial orientations.
Incorporating global rotation instead of local, canonicalized rotation is essential for generating motions such as squatting and crawling,
where the notion of heading is ill-defined and affects the quality of motion planning.
We refer readers to \cite{meng2024rethinking} and \cite{meng2025absolute} for similar insights and additional discussion.

\paragraph{Generative Neural Backbone in Latent Space}
Planning was conducted in the latent space,
where continuous motions are first encoded as a sequence of latent tokens as follows:
\begin{equation}
    \left\{z_t\right\}_{t=1}^{T/4} = \text{enc}\left(\left\{p_t, r_t\right\}_{t=1}^{T}\right),
\end{equation}
where $p_t$ and $r_t$ denote the pose configuration and root position at frame $t$, respectively.
In practice, the encoder operates with a downsampling rate of 4.
The latent token sequence is encoded by models such as Transformers or Conv1D networks to capture temporal consistency.

The in-betweening process in the token space is guided by two constraints: the starting and target keyframes,
denoted as $\left\{p_t, r_t\right\}_{t=1}^{4}$ and $\left\{p_t, r_t\right\}_{t=T-4}^{T}$ respectively.
Rather than training the network to predict the entire sequence of tokens from these sparse constraints in a single pass,
we adopted a masked token prediction approach \citep{yu2023magvit, luo2024open, guo2024momask, pinyoanuntapong2024mmm}.
In this framework, the neural backbone iteratively predicts and finalizes the subset of tokens for which it has the highest confidence,
progressively refining the prediction:
\begin{align}
     & h = \mathcal{F}\left(\left\{p_t, r_t\right\}_{t=1}^{4}, \left\{p_t, r_t\right\}_{t=T-4}^{T}, \left\{z_t\right\}_{t=1}^{T / 4}\right), \\
     & \mbox{Prob}(z_t) = \sigma(h).
\end{align}
This process is iterative, in which $\mathcal{F}(\cdot)$ denotes the neural backbone,
and $h$ represents the logits for each token position.
Token probabilities are computed by applying a softmax function $\sigma(\cdot)$ to the logits.
At the first iteration, all latent tokens are unknown, and we initialized the latent embedding with a learnable mask embedding,
$z_{\text{masked}}$.
During training, the proportion of masked tokens is uniformly sampled from the range $[100\%, 0\%]$.
During inference, a cosine schedule determines the proportion of tokens to finalize at each iteration,
specifically $1.0 - \cos\left(\frac{\pi}{2} \cdot \frac{L}{L_{\max}}\right)$, where $L$ is the current iteration and $L_{\max}$ is the maximum number of iterations.
After finalization of all tokens, the predicted tokens are used to reconstruct the kinematic motions and generate the robot control signals.

\paragraph{Root trajectory spring model}
We proposed to use an intuitive critically damped spring model to generate the root position and heading of the keyframes from user commands as follows:
\begin{equation}\label{equation:spring_model}
    x(t)=\left(x_T - x_0 + \left(v_0 + \frac{c}{2} \left(x_T - x_0\right)\right) t\right)e^{-\frac{c}{2} t},
\end{equation}
where $x_T$ denotes the target value, $x_0$ the initial value, $v_0$ the initial velocity, and $c$ the damping coefficient.
We applied this critically damped spring model to three quantities:
the pelvis position along the x-axis, 
the pelvis position along the y-axis, and 
the projected heading angle of the pelvis.
Damping coefficients of $5\ln(2)$ and $20\ln(2)$ were used for position and heading respectively.
The target values may be obtained directly from the controllers.
Alternatively, if the controller only specifies a desired velocity, we can compute the expected target positions after 1.0\,s using the desired velocity.
The target keyframes were then placed at the position and heading with $x(1.0)$, as computed by the spring model in \Cref{equation:spring_model}.
In practice, the planner is robust to the choice of damping coefficients---owing to its variable-length generation (0.8--2.4\,s) and strong in-betweening, the spring model can often be omitted entirely---but including it improves behavioral predictability and safeguards against unrealistic commands, such as abruptly reversing direction from 6.0\,m/s to $-$6.0\,m/s.

\paragraph{Keyframe Module and Application Integration}
Unlike traditional planners that rely on complex target keyframe generation (such as detailed footstep planning), our system specifies keyframes intuitively and with limited manual effort: navigation keyframes are placed along the target root trajectory using style clips, entertainment skills (such as boxing) use the most expressive segment of a matching clip, and interactive manipulation skills (such as squatting and kneeling) retrieve height-conditioned keyframes online. Combined with autoregressive replanning and robust variable-length in-betweening, this lets the planner generate the full distribution of transitional motions for a given skill from only a single clip, generalizing across styles such as walking, running, and crawling. Full per-mode keyframe specification is provided in \secref{app:planner_details}.
\subsection{Multi-modal Motion Generation}
\label{sec:method:genmo}
For multi-modal control (video, text, music), we adopted GEM~\citep{genmo2025}, a unified generalist model that handles both motion estimation and generation by treating estimation as constrained generation. GEM accepts mixed, time-varying conditions (text, audio, video) and produces human motion sequences via a diffusion-based prior. We integrated GEM with our system using sliding windows with overlap and inpainting-based transitions for low-latency generation. The generated human motions were fed into \shortname via the human motion encoder $\bs{\mathcal{E}}_h$.

\subsection{Deployment}
\label{sec:deployment}

Experiments were conducted on a Unitree G1 platform (29 actuated joints). All inference ran \emph{onboard} a Jetson Orin GPU using TensorRT with CUDA Graph acceleration, yielding 1--2\,ms per policy forward pass and ${\sim}$12\,ms for motion generation. The system used a multi-rate architecture with four concurrent loops: policy inference at 50\,Hz, command streaming at 500\,Hz, operator input at 100\,Hz, and kinematic planning at 10\,Hz. The encoder-decoder design allowed seamless switching between input interfaces (keyboard, gamepad, VR, network streams) by changing the active encoder, with no retraining required. All real-world experiments deployed the largest model (42M parameters). Full deployment details, including the multi-rate architecture, observation gathering pipeline, safety mechanisms, and usage modes, are provided in the Supplementary Materials (\secref{app:deployment}, \Cref{fig:deploy_pipeline}). Code is available at \url{https://github.com/NVlabs/GR00T-WholeBodyControl}.

\subsection{Validation of Key Design Choices}
\label{sec:ablations}

In this section, we validated key design choices through ablations on the test-content (out-of-distribution) and test-repetition splits, including quantizer design and encoder comparison (\Cref{tab:ablations}), a VLA action space comparison (\Cref{tab:vla_tasks}(B)), latent space alignment analysis (\Cref{fig:latent_alignment}), and kinematic planner validation.

\paragraph{FSQ Tokens vs.\ Explicit Poses for VLA}
A key motivation for quantization is downstream VLA learning. We compared two action spaces (\Cref{tab:vla_tasks}(B)): the VLA predicts FSQ tokens (78-dim: 64-dim token + 14-dim hands), decoded by the universal control policy, vs.\ the VLA directly predicts SMPL whole-body poses and hand joints (81-dim total). FSQ tokens outperformed SMPL by +42 percentage points on average (68\% vs.\ 27\%), with the gap widening on complex tasks (60\% vs.\ 0\% on soda-can-to-trash-can). We attributed this to the compactness of the quantized latent space: FSQ tokens provide a low-dimensional, discrete action space that is easier for the VLA to learn from teleoperated demonstrations, whereas the high-dimensional continuous SMPL pose space amplifies small prediction errors into large tracking failures.

\paragraph{Quantizer Design and Configuration}
We chose Finite Scalar Quantization (FSQ)~\citep{mentzer2023finite} over VQ-VAE~\citep{van2017neural} because FSQ avoids codebook collapse, a failure mode where large portions of a learned codebook are unused. Under our diverse motion distribution (33 categories, 8,447 sub-categories), this can be a major concern. For a fair comparison, we used a multi-head VQ-VAE with comparable capacity (4 heads, codebook size 512, 2 tokens). As shown in \Cref{tab:ablations}(A), FSQ outperformed VQ-VAE by 8.7\,mm MPJPE-L on test-content. We also studied the effect of quantizer capacity (\Cref{tab:ablations}(B)) by varying per-token levels and dimensions (all configurations use two tokens). Due to compute constraints, this sweep was run on 32 GPUs rather than 128. For example, FSQ-16-16 denotes 16 quantization levels and 16 dimensions per token. Increasing capacity consistently improved performance, with token dimension having a larger effect than quantization levels, suggesting that representational capacity matters more than quantization granularity for diverse motion tracking. We used FSQ-32-32 as our default configuration throughout the paper.

\begin{table}[htbp]
    \centering
    \footnotesize
    \setlength{\tabcolsep}{3.5pt}
    \begin{tabular}{l cccc c cccc}
        \toprule
                                                      & \multicolumn{4}{c}{\textbf{Test-Content (OOD)}} &                                    & \multicolumn{4}{c}{\textbf{Test-Repetition}}                                                                                                                                                                                             \\
        \cmidrule(lr){2-5} \cmidrule(lr){7-10}
        \textbf{Configuration}                        & SR\,(\%)                                        & MPJPE-L                            & Vel.\ Dist.                                  & Accel.\ Dist.                      &  & SR\,(\%)                           & MPJPE-L                            & Vel.\ Dist.                        & Accel.\ Dist.                      \\
        \midrule
        \multicolumn{10}{l}{\textit{(A) Quantizer design (128 GPUs)}}                                                                                                                                                                                                                                                                                              \\
        FSQ (ours)               & \textbf{99.3}              & \textbf{26.6} & \textbf{3.14}           & \textbf{1.17} &  & \textbf{99.6} & \textbf{25.5} & \textbf{3.22} & \textbf{1.23} \\
        VQ-VAE                   & 98.7                       & 35.3          & 3.76                    & 1.37          &  & 99.3          & 32.2          & 3.83          & 1.44          \\
        \midrule
        \multicolumn{10}{l}{\textit{(B) FSQ configurations (32 GPUs, 2 tokens each)}}                                                                                                                                                                                                                                                                              \\
        FSQ-16-16                & 96.9                       & 35.7          & 3.68                    & 1.26          &  & 97.5          & 32.7          & 3.75          & 1.31          \\
        FSQ-16-32                & 98.3                       & 29.7          & 3.39                    & 1.21          &  & 98.7          & 28.4          & 3.48          & 1.27          \\
        FSQ-32-16                & 98.3                       & 30.3          & 3.44                    & 1.22          &  & 98.4          & 28.9          & 3.52          & 1.28          \\
        FSQ-32-32 (ours)         & \textbf{98.8}              & \textbf{27.5} & \textbf{3.25}           & \textbf{1.19} &  & \textbf{99.3} & \textbf{26.3} & \textbf{3.34} & \textbf{1.25} \\
        \midrule
        \multicolumn{10}{l}{\textit{(C) Encoder comparison (128 GPUs)}}                                                                                                                                                                                                                                                                                            \\
        Robot ($\mathcal{E}_r$)  & \textbf{99.6}              & \textbf{23.8} & \textbf{2.89}           & \textbf{1.12} &  & 99.8          & \textbf{22.5} & \textbf{2.96} & \textbf{1.18} \\
        Human ($\mathcal{E}_h$)  & 99.6                       & 24.4          & 3.04                    & 1.24          &  & \textbf{99.8} & 23.1          & 3.11          & 1.30          \\
        Hybrid ($\mathcal{E}_m$) & 99.2                       & 26.5          & 3.25                    & 1.22          &  & 99.7          & 25.2          & 3.31          & 1.28          \\
        \bottomrule
    \end{tabular}
    \caption{\textbf{Ablation studies.} SR denotes success rate. Each entry reports a single evaluation per configuration on the full test split (descriptive; no statistical test applied). \textbf{(A)} FSQ outperforms VQ-VAE by 8.7\,mm MPJPE-L on test-content. \textbf{(B)} Higher quantizer capacity improves performance; token dimension matters more than levels. \textbf{(C)} All encoders maintain $>$99.2\% success; the human encoder shows only a +0.6\,mm MPJPE-L gap from the robot encoder.}
    \label{tab:ablations}
\end{table}

\paragraph{Multi-Encoder Performance and Consistency Losses}
Our multi-encoder design maps three heterogeneous input types (robot motion, human SMPL poses, and hybrid teleop commands) into a shared token space, aligned by the consistency losses $\mathcal{L}_{\text{token}}$ and $\mathcal{L}_{\text{cycle}}$. As shown in \Cref{tab:ablations}(C), all three encoders maintained $>$99.2\% success, with the human encoder showing only a +0.6\,mm MPJPE-L gap from the robot encoder despite operating on a different input format. The hybrid encoder showed a larger MPJPE-L (26.5\,mm, +2.7\,mm from the robot encoder) due to partial observability (only sparse upper-body keypoints). Removing the consistency losses caused an $8\times$ increase in cross-encoder divergence (\Cref{fig:latent_alignment}), confirming they are necessary for cross-encoder alignment. This alignment is critical for downstream VLA learning: since the VLA directly predicts tokens, teleoperation data collected via different encoders (human, hybrid, or robot) should occupy the same latent space to provide the VLA with a consistent training distribution.

\paragraph{Role of the Kinematic Motion Planner}
The kinematic planner (\secref{sec:method:planner}) is an application layer that converts high-level user intent into short-horizon kinematic references. Although the tracker is source-agnostic and compatible with alternative planners, ours unifies 25+ distinct skills and styles with a single real-time generative model, each requiring only one representative motion clip and no retraining. The tracker was independently validated on pre-recorded reference motions (\secref{sec:res:tracking}), and its robustness extends to planner-generated references through domain randomization on motion commands during training and the spring model that filters unrealistic commands at deployment.

\subsection{Statistical Analysis}
\label{sec:stats}

All statistical tests were two-sided, and $P$ values below $0.001$ are reported as $P<0.001$; exact values, test statistics, and the tabulated data underlying every panel of \Cref{fig:compare_baselines} are provided in Data~S1.

\textit{Error bars and uncertainty.} Shaded bands in the scaling curves (\Cref{fig:compare_baselines}, A to C) denote the mean $\pm$1 standard deviation (s.d.) over $n{=}6$ independent evaluation checkpoints per configuration. Bands in the velocity-tracking comparison (\Cref{fig:compare_baselines}, H to J) denote the mean $\pm$1 s.d.\ across repeated locomotion directions per commanded velocity. Success and survival rates are reported together with their underlying trial counts. The tracking metrics (MPJPE-L, velocity distance, and acceleration distance) are means over successfully tracked motions computed on 14 body links; because these metrics are undefined for early-terminated rollouts, no distributional test is applied to them.

\textit{Hypothesis tests.} Unpaired comparisons used Welch's two-sided $t$-test (unequal variances): scaling effects were assessed between the smallest and largest configuration on each axis, and success and survival rates were compared on per-motion (or per-run) binary outcomes, which at these sample sizes is equivalent to a two-proportion $z$-test. Matched comparisons used paired $t$-tests: velocity-tracking error against the specialist controller was paired across commanded velocities---reported separately for the low-velocity ($\leq$2\,m/s) and high-velocity ($>$2\,m/s) regimes, because the two methods cross over---and sim-versus-real success was paired across the 124 real-world evaluation motions. Real-world evaluation used one trial per motion sequence ($N{=}124$); VLA task success (\Cref{tab:vla_tasks}) was measured over $n{=}10$--$20$ trials per task with strict binary scoring. Ablation tables report a single evaluation per configuration and are therefore descriptive.

\section{Acknowledgments}

We thank Scott Reed, Wei Liu, You Liang Tan, Avnish Narayan, Fengyuan Hu, Yuqi Xie, Letian (Max) Fu, Mengda Xu, Ruijie Zheng, Davis Rempe, Xue Bin (Jason) Peng, Haotian Zhang, Yifeng Jiang, Anna Minx, John Malaska, Chen Tessler, Soha Pouya, Kaushil Prakashbhai Kundalia, Huihua Zhao, Xiaowei Jiang, Olivier Dionne, Michael De Ruyter, Michael Buttner, Qi Wang, Yeongho Seol, Mathis Petrovich, Sanja Fidler, Kaifeng Zhao, Spencer Huang, Gavriel State, Yurong ``Kelly'' Guo for their thoughtful discussions and help. We thank Amanpreet Singh, Leilee Naderi, Peter Pham, Rajeev Varma, Justin Lee and all the data collection team at GEAR for their support with data collection and providing feedback on teleoperation user experience. We thank Jeremy Chimienti and Tri Cao for their work ensuring robot readiness and conducting the necessary repairs. We also thank Tri Cao, Jazmin Sanchez, Demetria Quijada, and Jesse Yang for their help in filming and editing.

\paragraph{Author contributions} ZL, YY, TW, CL, FC contributed equally as co-first authors; they trained the policies, developed planners, designed the deployment pipelines, and integrated with the VLA. SC, ZC, JL, DM, QB, JP, DS, ZW, XD are core contributors to the framework, including system ID, VLA training, teleoperation, and model optimization. RD, CH, LS, EL, EJ, TH, HX, WX focused on retargeting, filtering the large-scale humanoid motion data, and codebase setup. SY, JK, YC, UI, LF, YZ provided leadership and guidance.

\paragraph{Funding} This research is funded by NVIDIA Corporation.

\paragraph{Competing interests} The authors are or were employees of NVIDIA Corporation at the time of this work. NVIDIA Corporation has filed a patent application related to this work.

Additional qualitative results and videos are available at the project website: \url{https://nvlabs.github.io/GEAR-SONIC/}.

\paragraph{Data, Code and Materials availability} The deployment and training code is available at \url{https://github.com/NVlabs/GR00T-WholeBodyControl}. A substantial portion of the motion-capture dataset has been publicly released as the BONES-SEED dataset, available at \url{https://huggingface.co/datasets/bones-studio/seed} (142,220 annotated motion sequences from 522 actors). An archival snapshot of the code, together with the tabulated data underlying all figures (Data~S1), is deposited at Zenodo (DOI: 10.5281/zenodo.21273312). No new materials were generated in this study. The paper, Supplementary Materials, archived code, and BONES-SEED release provide the information and data necessary to assess the reported conclusions.


\clearpage 


\bibliographystyle{unsrtnat}
\bibliography{science_template}

\appendix
\newpage
\clearpage
\begin{center}
    {\large\bfseries Supplementary Materials for}\\[0.5em]
    {\large\bfseries Supersizing Motion Tracking for Natural Humanoid Whole-Body Control}\\[1em]
    \parbox{0.9\textwidth}{\centering\small
        Zhengyi~Luo$^{\dagger}$, Ye~Yuan$^{\dagger}$, Tingwu~Wang$^{\dagger}$, Chenran~Li$^{\dagger}$, Fernando~Casta\~neda$^{\dagger}$, Sirui~Chen$^{\ast}$, Zi-Ang~Cao$^{\ast}$, Jiefeng~Li$^{\ast}$, David~Minor$^{\ast}$, Qingwei~Ben$^{\ast}$, Jinhyung~Park$^{\ast}$, David~Sami$^{\ast}$, Zi~Wang$^{\ast}$, Xingye~Da$^{\ast}$, Runyu~Ding, Cyrus~Hogg, Lina~Song, Edy~Lim, Eugene~Jeong, Tairan~He, Haoru~Xue, Wenli~Xiao, Simon~Yuen, Jan~Kautz, Yan~Chang, Umar~Iqbal, Linxi~``Jim''~Fan$^{\ddagger}$, Yuke~Zhu$^{\ddagger}$\par}\\[0.8em]
    {\small NVIDIA, 2788 San Tomas Expressway, Santa Clara, CA, USA}\\[0.3em]
    {\small $^{\dagger}$Co-first Authors,\quad $^{\ast}$Core Contributors,\quad $^{\ddagger}$Equal Advising}
\end{center}
\vspace{1.5em}

\renewcommand{\thefigure}{S\arabic{figure}}
\setcounter{figure}{0}
\renewcommand{\thetable}{S\arabic{table}}
\setcounter{table}{0}
\renewcommand{\thesubsection}{S\arabic{subsection}}
\setcounter{subsection}{0}
\renewcommand{\thesubsubsection}{\thesubsection.\arabic{subsubsection}}

\etocsettocstyle{\noindent\textbf{Contents}\vspace{2mm}}{}
\etocsetnexttocdepth{subsection}
\localtableofcontents

\subsection{Supplementary Methods}

\subsubsection{Implementation Details} \label{app:tables}

This section provides details for the network architecture, training hyperparameters, reward definitions, and domain randomization settings.

\paragraph{Network Architecture}
Table~\ref{tab:policy_params} reports the full configuration of the encoder-decoder architecture used by \shortname, including the robot-motion, human-motion, and hybrid encoders; the universal token quantizer parameters (FSQ levels and token dimension); the robot-control and auxiliary robot-motion decoders; and layer dimensions, MLP depth, activation functions, and latent sizes.

\paragraph{Training Hyperparameters}
Table~\ref{tab:ppo_params} lists all training configurations, including PPO hyperparameters, rollout length, and discount factors.

\paragraph{Reward Function}
Table~\ref{tab:rl_rewards} provides a complete breakdown of the reward terms used for motion tracking, including root position and orientation matching; body-link position, orientation, and velocity terms; end-effector position tracking; and action rate, joint limit, contact, anti-shake, and feet acceleration penalties.

\paragraph{Domain Randomization}
We randomized physical parameters including friction coefficients ($\mu_s$, $\mu_d$), the restitution coefficient ($e$), first-frame joint positions ($\bs{q}_0$), and the base center-of-mass position, periodically perturbed the robot's root linear and angular velocities to simulate external pushes, and perturbed the target motion commands $\goalstate$ during training. Table~\ref{tab:rand_runtime_eq} reports the ranges and distributions for all randomization parameters.

\paragraph{Universal Token Quantization}
We used Finite Scalar Quantization (FSQ)~\citep{mentzer2023finite} rather than a vector-quantized variational autoencoder (VQ-VAE)~\citep{van2017neural} to produce the universal token. FSQ avoids codebook collapse (a failure mode where large portions of the codebook go unused), requires no auxiliary commitment loss or codebook exponential-moving-average (EMA) updates, and provides clean straight-through gradient estimation that is compatible with joint PPO optimization. We validated these design choices in \secref{sec:ablations}, including FSQ vs.\ VQ-VAE, quantizer configuration (levels and dimensions), and multi-encoder alignment.

\begin{table}[t]
    \centering
    \footnotesize
    \setlength{\tabcolsep}{6pt}
    \begin{tabular}{l l l}
        \toprule
        \textbf{Module}         & \textbf{Architecture}                            & \textbf{Dims}                                            \\
        \midrule
        \multicolumn{3}{l}{\textit{Network configuration}}                                                                                    \\
        \quad Quantizer         & FSQ                                              & token dimensions $=D_z$; quantization levels $=L_z$      \\
        \quad Encoder (g1)      & MLP                                              & hidden$=[2048,1024,512,512]$                             \\
        \quad Encoder (teleop)  & MLP                                              & hidden $=[2048,1024,512,512]$                            \\
        \quad Encoder (smpl)    & MLP                                              & hidden $=[2048,1024,512,512]$                            \\
        \quad Decoder (actions) & MLP                                              & hidden $=[4096, 4096, 2048, 2048, 1024, 1024, 512, 512]$ \\
        \quad Decoder (refs)    & MLP                                              & hidden $=[2048,1024,512,512]$                            \\
        \quad Action dimension  & Diagonal Gaussian                                & $29$                                                     \\
        \quad Critic            & MLP                                              & hidden $=[4096, 4096, 2048, 2048, 1024, 1024, 512, 512]$ \\
        \multicolumn{2}{l}{\textit{Motion command}}                                                                                           \\
        \quad Future frames     & $F_r=F_h=F_m=10$ frames                                                                                     \\
        \quad Frame interval    & $\Delta t_r=\Delta t_m=0.1s$, $\Delta t_h=0.02s$                                                            \\
        \bottomrule
    \end{tabular}
    \caption{Universal control policy architecture and hyperparameters.}
    \label{tab:policy_params}
\end{table}
\begin{table}[t]
\centering
\footnotesize
\setlength{\tabcolsep}{6pt}
\begin{tabular}{l l}
\toprule
\textbf{Training hyperparameter} & \textbf{Value} \\
\midrule

Num parallel envs per GPU & $4096$ \\
Num steps per env & 24 \\
Learning epochs  & $5$ \\
Num mini-batches & $4$ \\
Discount $\gamma$ & $0.99$ \\
GAE $\lambda$ & $0.95$ \\
Clip parameter & $0.2$ \\
Entropy coefficient & $0.013$ \\
Value loss coefficient & $1.0$ \\
Actor learning rate & $2{\times}10^{-5}$\\
Critic learning rate & $1{\times}10^{-3}$ \\
Max gradient norm & $0.1$ \\
Desired KL & $0.01$ \\
Adaptive LR min/max & $[1{\times}10^{-5}$ , $2{\times}10^{-4}]$ \\
Init noise std & $0.05$ \\
Actor std clamp min/max & $[0.001, 0.5]$ \\
Adaptive sampling bin size & 1s \\
Adaptive sampling failure rate cap & $\beta = 200 $ \\
Adaptive sampling blending hyperparameter & $\alpha = 0.1$ \\
\bottomrule
\end{tabular}
\caption{Training hyperparameters.}
\label{tab:ppo_params}
\end{table}

\begin{table}[t]
    \centering
    \footnotesize
    \setlength{\tabcolsep}{6pt}
    \begin{tabular}{l l c}
        \toprule
        \textbf{Reward term}                              & \textbf{Equation}                                                                                                                                                           & \textbf{Weight}                \\
        \midrule
        \multicolumn{3}{l}{\textit{Tracking rewards $\rewardfunc(\selfstate, \goalstate)$}}                                                                                                                                                                              \\
        \quad Root position          & $r^{\text{root}}_{\text{pos}}(t)=\exp\!\big(- \|\bs{p}^p_{t,r}-\bs{p}^g_{t,r}\|_2^2 / 0.3^2\big)$                                                      & 0.5       \\
        \quad Root orientation                            & $r^{\text{root}}_{\text{ori}}(t)=\exp\!\big(- \|\bs{o}^p_{t,r}-\bs{o}^g_{t,r}\|_2^2 / 0.4^2\big)$                                                                           & 0.5                            \\
        \quad Body link pos (rel.)                        & $r^{\text{body}}_{\text{pos}}(t)=\exp\!\Big(-\tfrac{1}{|\mathcal{B}|}\!\sum_{b\in\mathcal{B}}\|\bs{p}^{p,\text{rel}}_{t,b}-\bs{p}^{g,\text{rel}}_{t,b}\|_2^2 / 0.3^2\Big)$  & 1.0                            \\
        \quad Body link ori (rel.)                        & $r^{\text{body}}_{\text{ori}}(t)=\exp\!\Big(-\tfrac{1}{|\mathcal{B}|}\!\sum_{b\in\mathcal{B}} \|\bs{o}^{p,\text{rel}}_{t,b}-\bs{o}^{g,\text{rel}}_{t,b}\|_2^2 / 0.4^2\Big)$ & 1.0                            \\
        \quad Body link lin.\ vel                         & $r^{\text{body}}_{\text{lin}}(t)=\exp\!\Big(-\tfrac{1}{|\mathcal{B}|}\!\sum_{b\in\mathcal{B}}\|\bs{v}^p_{t,b}-\bs{v}^g_{t,b}\|_2^2 / 1.0^2\Big)$                            & 1.0                            \\
        \quad Body link ang.\ vel                         & $r^{\text{body}}_{\text{ang}}(t)=\exp\!\Big(-\tfrac{1}{|\mathcal{B}|}\!\sum_{b\in\mathcal{B}}\|\bs{\omega}^p_{t,b}-\bs{\omega}^g_{t,b}\|_2^2 / 3.14^2\Big)$                 & 1.0                            \\
        \quad End-effector position       & $r^{\text{ee}}_{\text{pos}}(t)=\exp\!\big(-\tfrac{1}{5}\!\sum_{k\in\mathcal{K}}\|\bs{p}^p_{t,k}-\bs{p}^g_{t,k}\|_2^2 / 0.1^2\big)$                     & 2.0       \\
        \midrule
        \multicolumn{3}{l}{\textit{Penalty terms $\mathcal{P}(\selfstate, \bs{a}_t)$}}                                                                                                                                                                                   \\
        \quad Action rate                                 & $r_{\text{act}}(t)=\|\bs{a}_t-\bs{a}_{t-1}\|_2^2$                                                                                                                           & $-$0.1                         \\
        \quad Joint limit                                 & $r_{\text{jlim}}(t)=\sum_{j} \mathbbm{1}[\bs{q}_{t,j} \notin [\bs{q}_{t,j}^{\text{min}}, \bs{q}_{t,j}^{\text{max}}]]$                                                       & $-$10.0                        \\
        \quad Undesired contacts                          & $r_{\text{contact}}(t)=\sum_{c \notin \{\text{ankles, wrists}\}} \mathbbm{1}[\|\bs{F}_c\|>1.0\text{N}]$                                                                     & $-$0.1                         \\
        \quad Anti-shake (ang.\ vel) & $r_{\text{shake}}(t)=\sum_{k\in\{\text{wrists, head}\}}\|\bs{\omega}_{t,k}\|_2^2 \cdot \mathbbm{1}[\|\bs{\omega}_{t,k}\|>1.5]$                         & $-$5e-3   \\
        \quad Feet acceleration      & $r_{\text{feet}}(t)=\sum_{k\in\{\text{ankles}\}}\|\ddot{\bs{q}}_{t,k}\|_2^2$                                                                           & $-$2.5e-6 \\
        \bottomrule
    \end{tabular}
    \caption{Reward design. Superscript $g$: goal/target; $p$: current state; $\mathcal{B}$: tracked body links; $\mathcal{K}$: VR keypoints (head, both wrists, both ankles); ``rel'': relative to root frame. }
    \label{tab:rl_rewards}
\end{table}

\begin{table}[t]
\centering
\footnotesize
\setlength{\tabcolsep}{6pt}
\begin{tabular}{l l}
\toprule
\textbf{Domain Randomization} & \textbf{Sampling Distribution} \\
\midrule
\multicolumn{2}{l}{\textit{Physical parameters}} \\
\quad Static friction coefficients & $\mu_s \sim \mathcal{U}[0.3,1.6]$ \\
\quad Dynamic friction coefficients & $\mu_d \sim \mathcal{U}[0.3,1.2]$ \\
\quad Restitution coefficient & $e \sim \mathcal{U}[0,0.5]$ \\
\quad Default joint positions & $\bs{q}_0 \sim \bs{q}_0 + \mathcal{U}[-0.01, 0.01]$ \\
\quad Base COM offset (x, y, z) & $\Delta x \sim \mathcal{U}[-0.075,0.075],\ \Delta y \sim \mathcal{U}[-0.1,0.1],\ \Delta z \sim \mathcal{U}[-0.1,0.1]$ \\
\multicolumn{2}{l}{\textit{Root velocity perturbations (external pushes)}} \\
\quad Root linear vel (x, y, z) & $v_x \sim \mathcal{U}[-0.5,0.5],\ v_y \sim \mathcal{U}[-0.5,0.5],\ v_z \sim \mathcal{U}[-0.2,0.2]$ \\
\quad Push duration & $\Delta t \sim \mathcal{U}[1,3]\text{s}$ \\
\quad Root angular vel & $\omega_{\text{roll}} \sim \mathcal{U}[-0.52,0.52],\ \omega_{\text{pitch}} \sim \mathcal{U}[-0.52,0.52],\ \omega_{\text{yaw}} \sim \mathcal{U}[-0.78,0.78]$ \\
\multicolumn{2}{l}{\textit{Target motion perturbations ($\goalstate$)}} \\
\quad Target position jitter & $\Delta \bs{p}^g \sim \mathcal{U}[-0.05,0.05]^3$ (x,y: $\pm 0.05$, z: $\pm 0.01$) \\
\quad Target orientation jitter & $\Delta \phi_{\text{roll}}, \Delta \phi_{\text{pitch}} \sim \mathcal{U}[-0.1,0.1],\ \Delta \phi_{\text{yaw}} \sim \mathcal{U}[-0.2,0.2]$ \\
\quad Target linear vel jitter & $\Delta \bs{v}^g \sim \mathcal{U}[-0.5,0.5]^3$ (x,y: $\pm 0.5$, z: $\pm 0.2$) \\
\quad Target angular vel jitter & $\Delta \omega_{\text{roll}}, \Delta \omega_{\text{pitch}} \sim \mathcal{U}[-0.52,0.52],\ \Delta \omega_{\text{yaw}} \sim \mathcal{U}[-0.78,0.78]$ \\
\quad Target joint jitter & $\Delta \bs{q}_t^g \sim \mathcal{U}[-0.1,0.1]$ \\
\bottomrule
\end{tabular}
\caption{Domain randomization parameters applied during training. $\mathcal{U}[\cdot]$: uniform distribution.}
\label{tab:rand_runtime_eq}
\end{table}

\subsubsection{Deployment Architecture} \label{app:deployment}

Experiments were conducted on a Unitree G1 platform (29 actuated joints) using the built-in joint-level PD controller. The learned policy output desired joint angles that were passed directly to the PD interface. All components of the inference and management stack ran \emph{onboard}, leveraging the Jetson Orin GPU to minimize feedback latency and improve timing determinism.

\begin{figure}[t!]
    \centering
    \includegraphics[width=\textwidth]{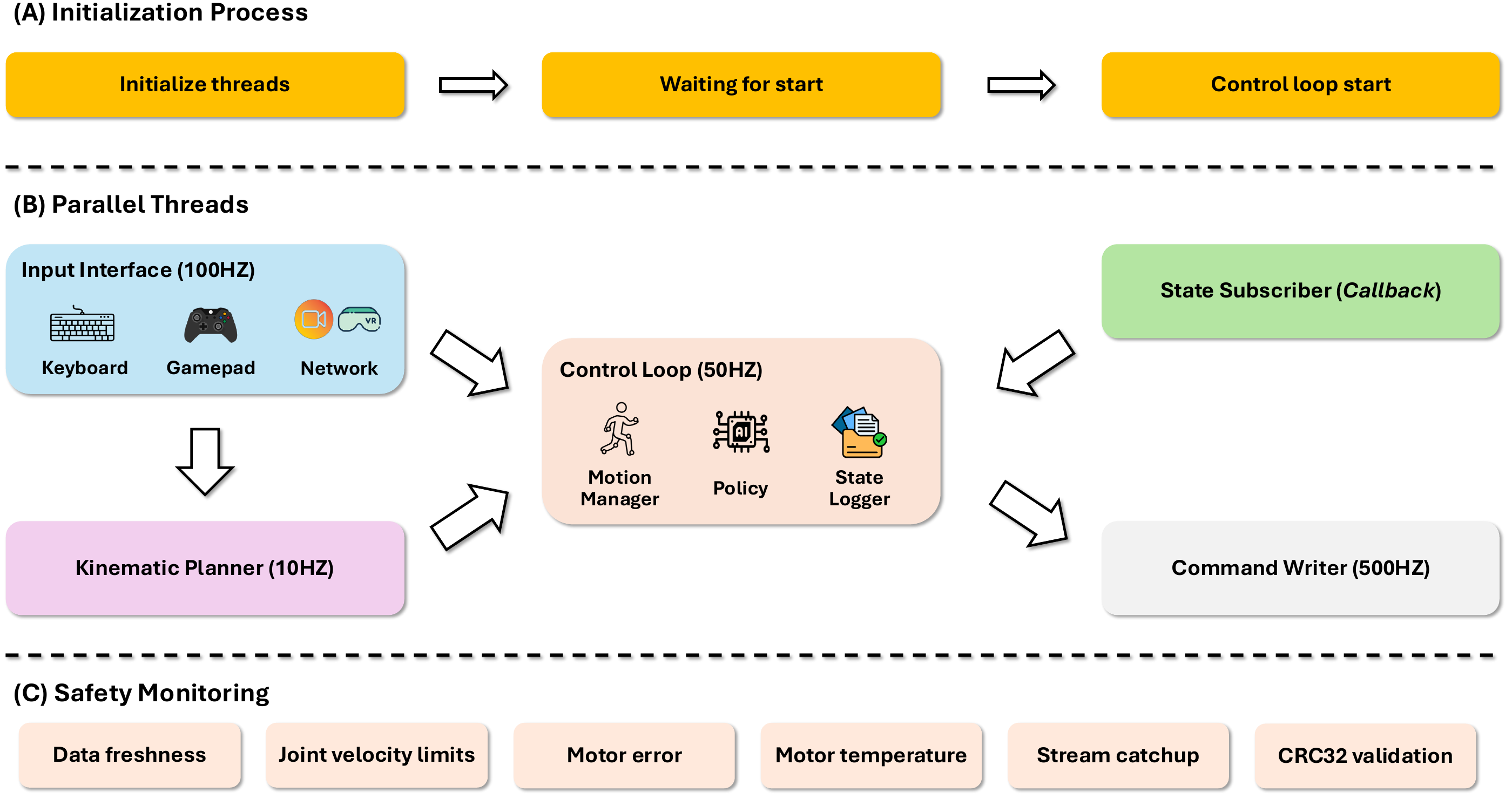}
    \caption{\textbf{Deployment architecture of \shortname.} All inference and management run onboard the Jetson Orin GPU. \textbf{(A)}~Initialization: worker threads are spawned, the system waits for the operator's start signal, and the control loop begins. \textbf{(B)}~Parallel threads run concurrently at matched rates and communicate under a latest-data-wins convention: an input interface (100\,Hz; keyboard, gamepad, VR, or network), a kinematic planner (10\,Hz), the control loop (50\,Hz; motion manager, policy, and state logger), a state subscriber callback, and a command writer (500\,Hz). \textbf{(C)}~Safety monitoring continuously checks data freshness, joint-velocity limits, motor error, motor temperature, stream catch-up, and CRC32 validation.}
    \label{fig:deploy_pipeline}
\end{figure}

\paragraph{Multi-rate architecture}
As shown in \Cref{fig:deploy_pipeline}, the system was organized into four concurrent loops, each running at a rate matched to its role. The \textbf{control loop (50\,Hz)} assembled observations from the recent robot state and reference motion, invoked the encoder and policy networks, and synthesized joint-space targets; its 50\,Hz rate matched the simulation step used during training. The \textbf{command writer (500\,Hz)} was a separate high-rate process that streamed motor targets through the Unitree low-level API, continuously publishing the latest commands without blocking the policy loop. The \textbf{input interface (100\,Hz)} sampled operator input independently, supporting seamless switching between keyboard, gamepad, VR controllers, or networked streams. The \textbf{kinematic planner (10\,Hz)}, when employed, proposed short-horizon reference trajectories based on the operator's high-level commands.
Each loop operated on consistent snapshots of the robot state, and we adopted a ``latest-data-wins'' convention; inter-thread buffers protected by reader-writer locks allowed each consumer to always read the most recent data without blocking producers, so transient delays did not stall the system.

\paragraph{Unified motion interface}
A central design choice was that the control loop was \emph{source-agnostic} with respect to the reference motion. All motion sources, including pre-recorded clips, planner-generated trajectories, and externally streamed data, populated the same \texttt{MotionSequence} structure, which stored per-frame joint positions, joint velocities, root poses, and optionally SMPL body data in flat arrays with frame-indexed accessors. A single shared pointer and a frame counter defined what the policy saw at each tick. Switching between sources was an atomic pointer swap under a mutex, with the frame counter reset to zero. Because the encoder mode was carried as a field on the motion sequence itself, changing the motion source automatically selected the appropriate encoder mode; for example, switching from a pre-recorded clip (mode 0, joint-based encoding) to an externally streamed SMPL sequence (mode 2, SMPL-based encoding) required no explicit mode management by the operator.

\paragraph{State logging and observation gathering}
At each control tick, the system first read IMU orientation, angular velocity, and joint positions/velocities from the hardware, remapped them to the policy's joint ordering, and pushed a timestamped snapshot into a dual-purpose state logger. The logger maintained both a fixed-capacity ring buffer for real-time access and per-signal CSV files for offline analysis. History-based observations (such as the past 10 frames of proprioceptive quantities) were retrieved from this ring buffer via strided lookback, with zero-padding when the buffer contained fewer entries than requested (for example, at startup).

The choice of observations to gather, in what order, and at what offsets was entirely driven by a YAML configuration file. At startup, each entry in the config was matched against a registry of named observation functions, and the system pre-computed a flat list of \texttt{(function, offset, dimension)} triples. The control loop then simply iterated this list, calling each function to fill its slice of the observation vector. This design allowed the same binary to serve different policy architectures by swapping a single config file, without recompilation.

\paragraph{GPU inference}
Both the interactive kinematic motion planner and the policy inference executed onboard the Jetson Orin GPU using TensorRT with CUDA Graph acceleration: a warmup pass recorded the complete GPU execution graph, and all subsequent calls replayed it with new data, yielding reliable low-variance execution, 1--2\,ms per forward pass for the policy and ${\sim}$12\,ms for motion generation.

\paragraph{Heading alignment}
When a new motion began, the system captured the robot's current yaw from the IMU and computed a yaw-only rotation that aligned the reference motion's initial heading to the robot's facing direction. This rotation was applied to all subsequent reference root poses, so the robot could execute any motion starting from its current heading. The operator could inject heading corrections at runtime to steer the robot.

\paragraph{Startup and safety}
The system followed a three-phase startup: it linearly interpolated the robot from its current configuration to a nominal standing pose over 3 seconds, held until the operator signaled readiness, then activated the policy loop. A joint velocity watchdog triggered an immediate stop if any joint exceeded a predefined threshold. If the encoder's required input data became unavailable during playback, the system automatically fell back to the next available encoder mode, maintaining continuous operation. Upon any stop event, all joints transitioned to a pure-damping mode ($K_p{=}0$, $K_d{=}8$\,Nm$\cdot$s/rad) that let the robot gently comply rather than holding position or going limp. All motor commands carried CRC32 checksums validated by the firmware before execution.

\paragraph{Data tracking}
Pre-recorded reference motions were loaded from CSV directories at startup. The operator selected a clip and triggered playback; the frame counter advanced at 50\,Hz, and the policy tracked the reference. When the clip ended, playback paused and the frame reset. This mode was used for evaluation and for motions that required precise, repeatable reference trajectories.

\paragraph{Motion planning}
When the planner was enabled, the planner thread generated full-body trajectories at 30\,fps temporal resolution (up to 64 frames, ${\sim}2$\,s of predicted motion), replanning at up to 10\,Hz, conditioned on a context window of 4 recent frames and operator commands (locomotion mode, target velocity, facing direction). At initialization, the planner context was set to a canonical pose regardless of the robot's actual orientation; subsequent replans built the context from the previously generated trajectory, so the planner autoregressively continued in the coordinate frame established at initialization. The 30\,Hz output was resampled to 50\,Hz using linear interpolation for positions and SLERP for quaternions, with velocities recovered by finite differences. When a new plan arrived, the system cross-faded from the old trajectory to the new one over 8 frames (160\,ms). The replanning interval adapted to the locomotion mode (for example, every tick for running and less frequently for walking), and mode or heading changes triggered immediate replanning.

\paragraph{Motion streaming}
External systems could stream motion data to the robot over ZMQ or ROS2. Incoming frames were merged into a sliding-window \texttt{MotionSequence} using a stream merger that handled variable-rate arrival and catch-up. The streamed motion could contain joint-level data, SMPL body poses, VR tracking targets, or any combination thereof. A special token-only streaming protocol allowed an external process to bypass the onboard encoder entirely, sending pre-computed latent tokens that were injected directly into the policy's observation vector. When streaming was active, the frame counter held at the latest available frame, waiting for new data to arrive.

Code is available at \url{https://github.com/NVlabs/GR00T-WholeBodyControl}.

\subsubsection{Kinematic Planner Details} \label{app:planner_details}

This section provides the full per-mode keyframe specification for the generative kinematic motion planner (\secref{sec:method:planner}).

\paragraph{Navigation control}
Target keyframes were generated by placing a randomly selected segment from the navigation clips of the desired style at the target root trajectory. Despite this simplicity, our model consistently produced natural and smooth motions that aligned well with the specified style. We attributed this to the model's flexible, variable-length motion generation and its robust in-betweening capabilities. Additionally, the autoregressive replanning ensured that the generated motion was continually refreshed before reaching the end of any given clip, thus minimizing dependence on the specific spatial details of the chosen target keyframes. This approach generalized to other motion styles such as walking, running, and crawling.

\paragraph{Entertainment tasks}
For entertainment tasks such as boxing, target keyframes were determined by selecting the most expressive segment (such as the frames with maximal arm extension for a punch) from motion clips that matched the desired style. We also supported motion layering, where the upper body was specified and the lower body was generated accordingly by the planner, enabling predefined behaviors.

\paragraph{Interactive manipulation modes}
For interactive modes needed in manipulation tasks, such as squatting or kneeling, keyframes were retrieved online from the motion clip library according to the desired height. Unlike traditional approaches that require an extensive motion library, our system needed only a single clip to generate the full distribution of transitional motions for a given skill.

\subsubsection{VR Teleoperation and VLA Integration Details} \label{app:teleop_vla}

This section provides full details for the teleoperation interfaces (\secref{sec:res:teleop}) and VLA integration (\secref{sec:vla_control}) summarized in the main text.

\paragraph{VR-Based Whole-Body Teleoperation}
We developed a full-body VR teleoperation system using the PICO whole-body motion-tracking interface~\citep{zhao2025xrobotoolkit}. This required wearing the PICO headset, two ankle trackers, and the handheld VR controllers. The PICO interface provided human motion as full-body human pose estimates in SMPL~\citep{SMPL:2015} format. Specifically, the VR interface sent 63-dimensional ($21 \times 3$) whole-body joint positions, a target root orientation as a 4-dimensional quaternion, 6-dimensional ($2 \times 3$) wrist joint angles, and 14-dimensional ($2 \times 7$) dexterous hand joint positions, yielding an 87-dimensional continuous stream per frame. The tracked human motion was streamed in real time to the universal control policy and encoded via the human motion encoder $\bs{\mathcal{E}}_h$ into the universal token space.

\paragraph{VR-Based 3-Point Teleoperation} To enable scalable and portable data collection for tasks that do not require precise foot placement, we introduced a lightweight mobile bimanual VR teleoperation interface that operated with the PICO headset and two handheld controllers (no ankle trackers needed). The teleoperation interface output a compact command consisting of three upper-body SE(3) poses (head and both wrists), hand joint angles, waist height, a locomotion mode (slow walk or fast walk), and a navigation command specifying the desired root velocity and heading. These signals were fed into the kinematic motion planner and hybrid encoder $\bs{\mathcal{E}}_m$. The generated motion was then tracked by the universal tracking policy.

\paragraph{3-Point VLA Integration}
Using the 3-point interface, we collected 300 teleoperated demonstrations of a mobile pick-and-place task: the robot walked to a randomly placed apple on a table, grasped it with the right hand, and placed it on a randomly positioned plate. We fine-tuned a GR00T N1.5 model~\citep{gr00t-n1_5-blog, bjorck2025gr00t} on these trajectories. The VLA output the same teleoperation-format control signals (three upper-body poses, base height, navigation command), which were fed into the kinematic planner and hybrid encoder and executed via the universal control policy. On this task, the system attained a 90\% success rate over 20 trials.

\paragraph{Whole-Body VLA Integration} For the whole-body tasks, the VLA predicted a 78-dimensional action comprising a 64-dimensional universal motion token and 14-dimensional hand joint angles. Task descriptions and success rates are provided in the main text (\secref{sec:vla_control}, \Cref{tab:vla_tasks}). Additional training details: the soda-can-to-trash-can task used both a multi-object dataset (1,000 trajectories including soda can, paper cup, cardboard, and other disposable items) and a single-object dataset (150 soda-can trajectories). The action space ablation (\Cref{tab:vla_tasks}(B)) compared FSQ tokens against explicit SMPL poses; we observed that SMPL predictions produced unsafe, jerky motions with poor directional control, whereas token predictions yielded substantially smoother behavior and higher task success rate.

\subsection{Supplementary Discussion}

\subsubsection{Motion Dataset Samples} \label{app:dataset_samples}

\begin{figure}[htbp]
    \centering
    \includegraphics[width=\textwidth]{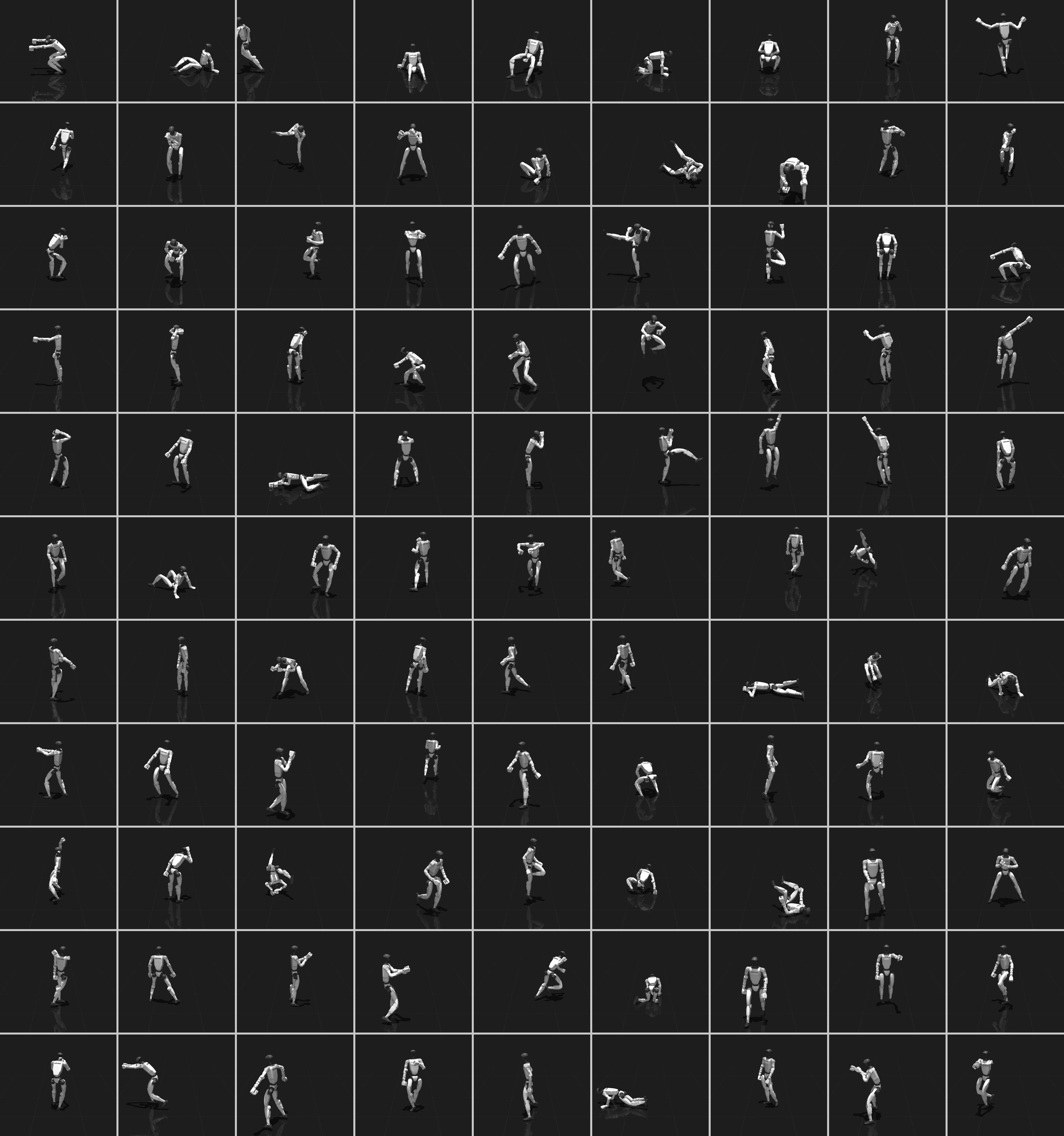}
    \caption{\textbf{Random samples from our large-scale humanoid motion dataset.} Each tile shows a frame from a distinct motion sequence retargeted to the Unitree G1 humanoid, illustrating the diversity of behaviors---including locomotion, dance, sports, combat, gestures, and object interactions---that span the 33 motion categories used for training.}
    \label{fig:dataset_panorama}
\end{figure}

\Cref{fig:dataset_panorama} shows random samples from our large-scale humanoid motion dataset, illustrating the diversity of behaviors used for training (see the \nameref{sec:method:data} section in the main text).

\subsubsection{Qualitative Analysis of Success and Failure Motions} \label{app:success_failure}

\begin{figure}[htbp]
    \centering
    \includegraphics[width=\textwidth]{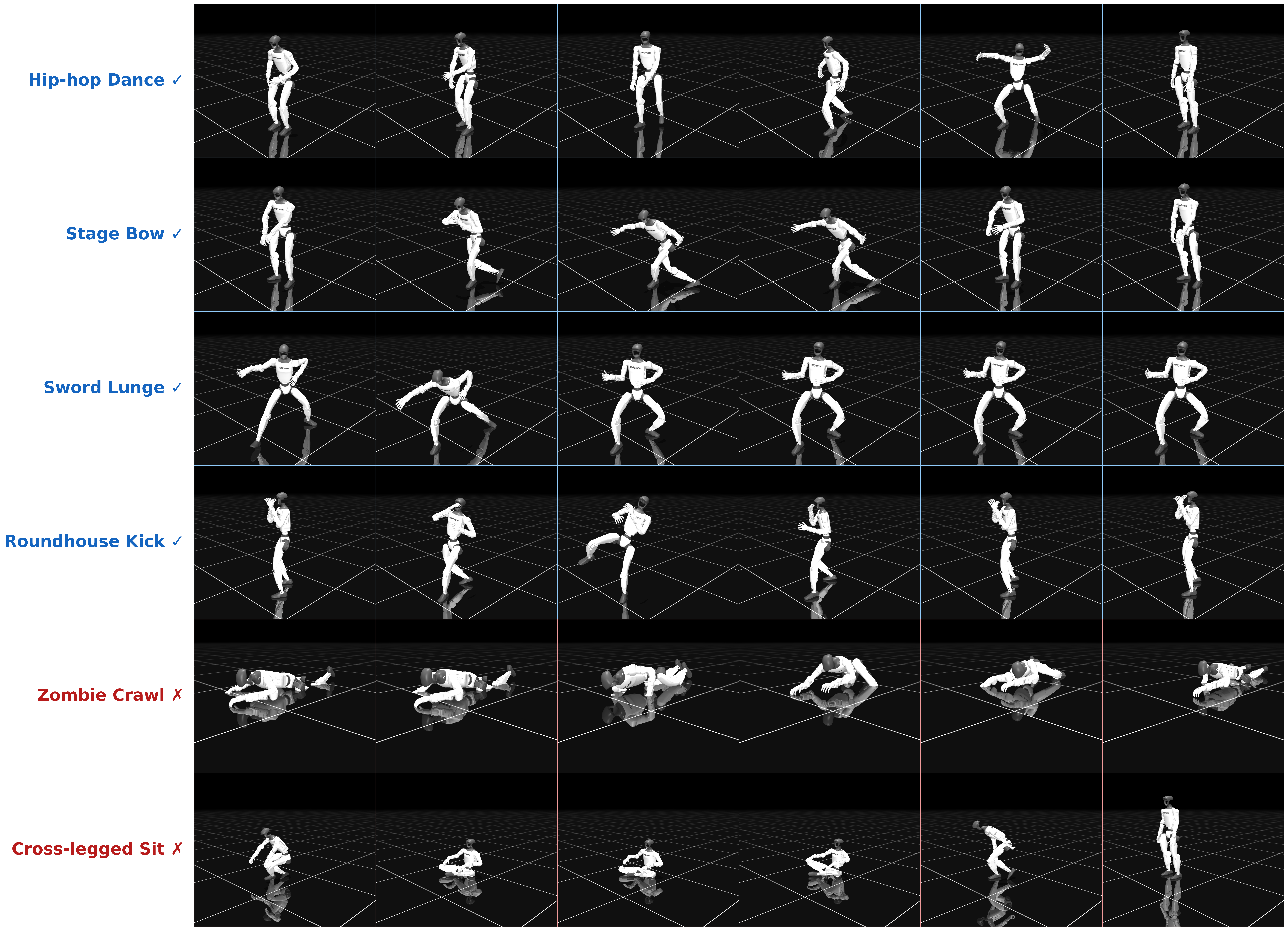}
    \caption{\textbf{Out-of-distribution motion sequences from the test-content split.} Unseen motions that are successfully tracked (hip-hop dance, stage bow, sword lunge, roundhouse kick) and unseen motions where tracking fails (zombie crawl, cross-legged sit).}
    \label{fig:success_failure}
\end{figure}

To complement the quantitative evaluation in \secref{sec:res:tracking}, we visualized out-of-distribution motion sequences from the test-content split to illustrate the range of unseen behaviors our tracker could handle and where it failed (\Cref{fig:success_failure}). The successfully tracked OOD motions---imitated despite never being encountered during training---include hip-hop dance, stage bow, sword lunge, and roundhouse kick. The failure cases include zombie crawl (an extreme floor-level motion that exceeds the humanoid's kinematic limits) and cross-legged sit (a static pose requiring sustained ground contact in a configuration far from the training distribution). These failures indicate that motions requiring sustained or complex ground contact remain challenging for the current system.

\subsubsection{Real-World Evaluation Motions} \label{app:real_eval_motions}

\begin{figure}[htbp]
    \centering
    \includegraphics[width=\textwidth]{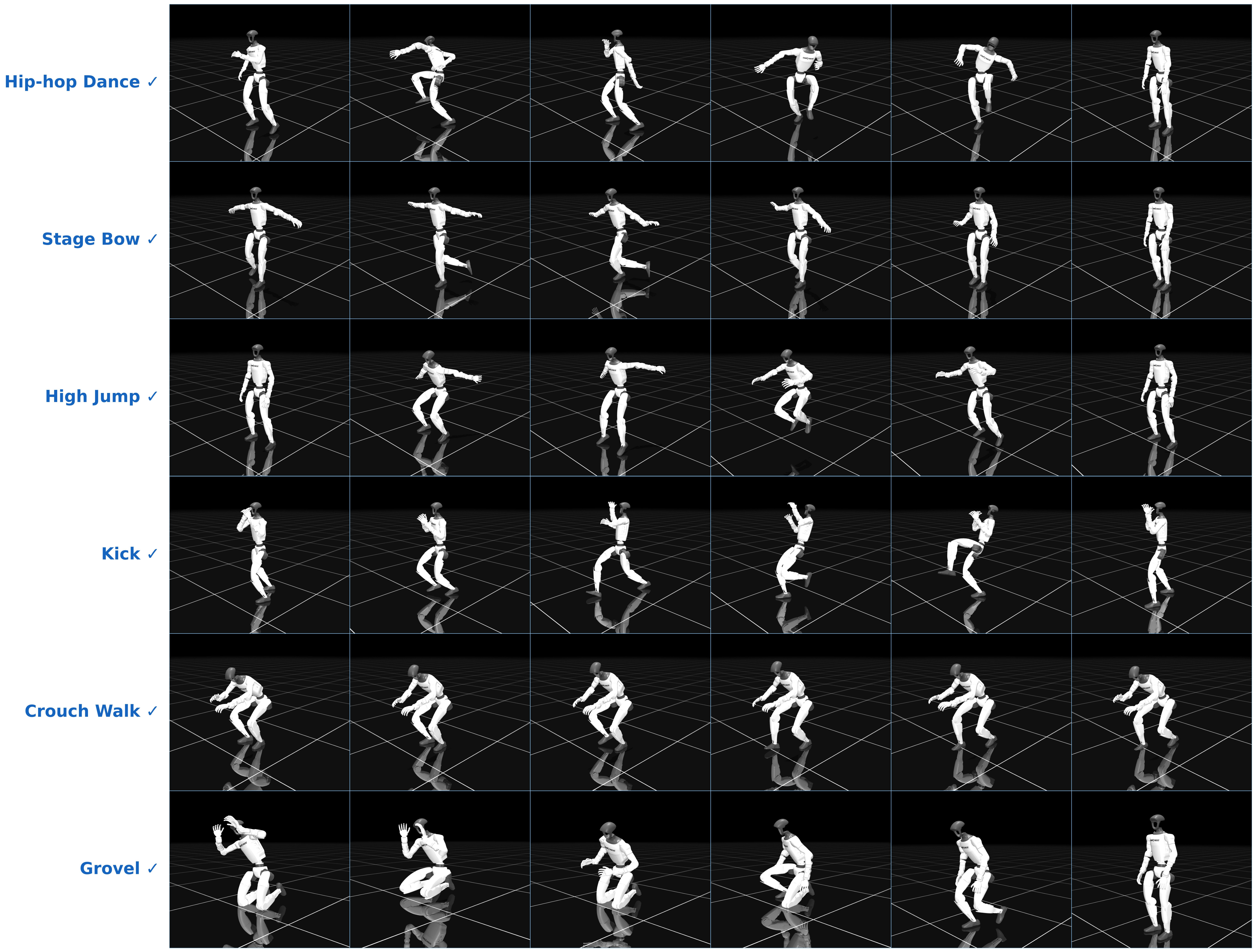}
    \caption{\textbf{Representative motions from the 124-sequence real-world evaluation set.} All shown motions are successfully tracked on the physical Unitree G1 robot.}
    \label{fig:real_eval_motions}
\end{figure}

We evaluated \shortname on 124 diverse motion sequences deployed on the real Unitree G1 robot. \Cref{fig:real_eval_motions} shows representative examples from the evaluation set, spanning hip-hop dance, stage bow, high jump, kick, crouch walk, and grovel. All motions shown were successfully tracked in the real world.

\subsubsection{Robustness to External Pushes} \label{app:robustness}
\begin{figure}[htbp]
    \centering
    \includegraphics[width=\textwidth]{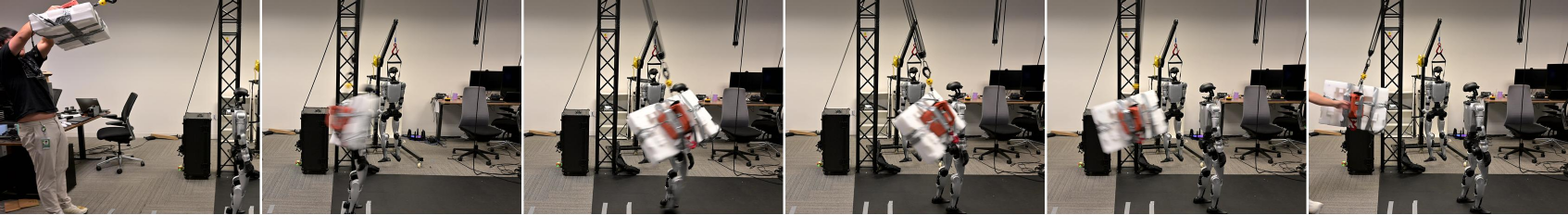}
    \caption{\textbf{Robustness test.} An approximately 11\,kg (25\,lb) object was dropped onto the robot from above head height during policy execution. The robot absorbed the impact, maintained balance, and continued tracking. No recovery module or policy adaptation was used.}
    \label{fig:appendix_push}
\end{figure}

To assess robustness against strong real-world disturbances, we dropped an approximately 11\,kg (25\,lb) object onto the robot from above head height while it executed the tracking policy (\Cref{fig:appendix_push}). The robot absorbed the impact, maintained balance, and continued motion tracking without any recovery module or policy adaptation.

\subsubsection{Scaling Analysis of Specialist Controllers} \label{app:homie_scaling}

\begin{figure}[htbp]
    \centering
    \includegraphics[width=0.8\textwidth]{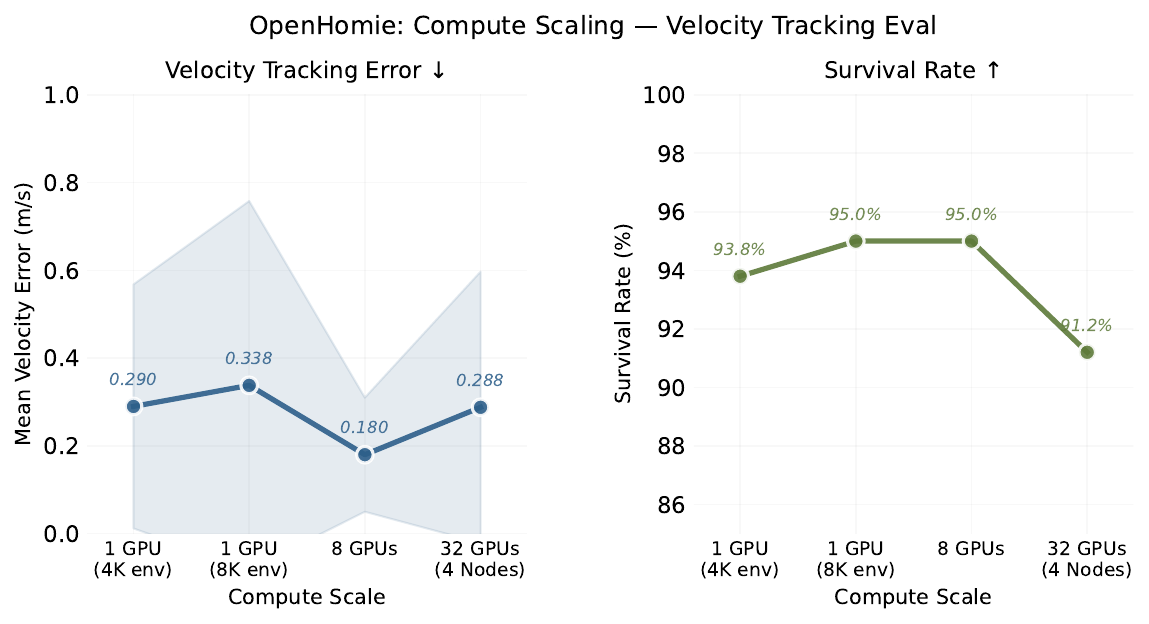}
    \caption{\textbf{OpenHomie velocity tracking evaluation vs.\ compute scale.} Mean velocity tracking error (m/s) and survival rate (\%), each shown as a function of training compute. Each configuration was evaluated over $n{=}80$ runs (20 commanded velocities from 0.1 to 2.0\,m/s $\times$ 4 directions, 5\,s each). On the tracking-error plot, the line denotes the mean and the shaded band mean $\pm$1 standard deviation (s.d.) over the $n{=}80$ runs; the survival rate is the percentage of those runs completed without a fall.}
    \label{fig:homie_scaling}
\end{figure}

To support the argument that motion tracking exhibits favorable scaling properties compared to specialist controllers (\secref{sec:res:tracking}), we studied the scaling behavior of OpenHomie~\citep{ben2025homie}, a state-of-the-art locomotion controller optimized for velocity tracking. \Cref{fig:homie_scaling} shows OpenHomie's velocity tracking error and survival rate as a function of compute scale, ranging from 1 GPU (4k environments) to 32 GPUs (4 nodes). OpenHomie's performance peaked at 8 GPUs (0.18\,m/s tracking error, 95.0\% survival) and did not improve when scaling to 32 GPUs (0.29\,m/s error, 91.2\% survival). This stood in contrast to \shortname, where scaling from 2 to 16 nodes yielded consistent improvements in both success rate and tracking accuracy (\Cref{fig:compare_baselines}A--C). We attributed this difference to the nature of the training objective: OpenHomie used task-specific reward engineering for locomotion, which saturated once the policy mastered the target behavior, whereas motion tracking provided dense per-frame supervision across a diverse motion distribution that continued to benefit from additional capacity and data throughput.

\subsubsection{Latent Space Alignment of the Multi-Encoder Design} \label{app:latent_alignment}
\begin{figure}[htbp]
    \centering
    \includegraphics[width=0.95\textwidth]{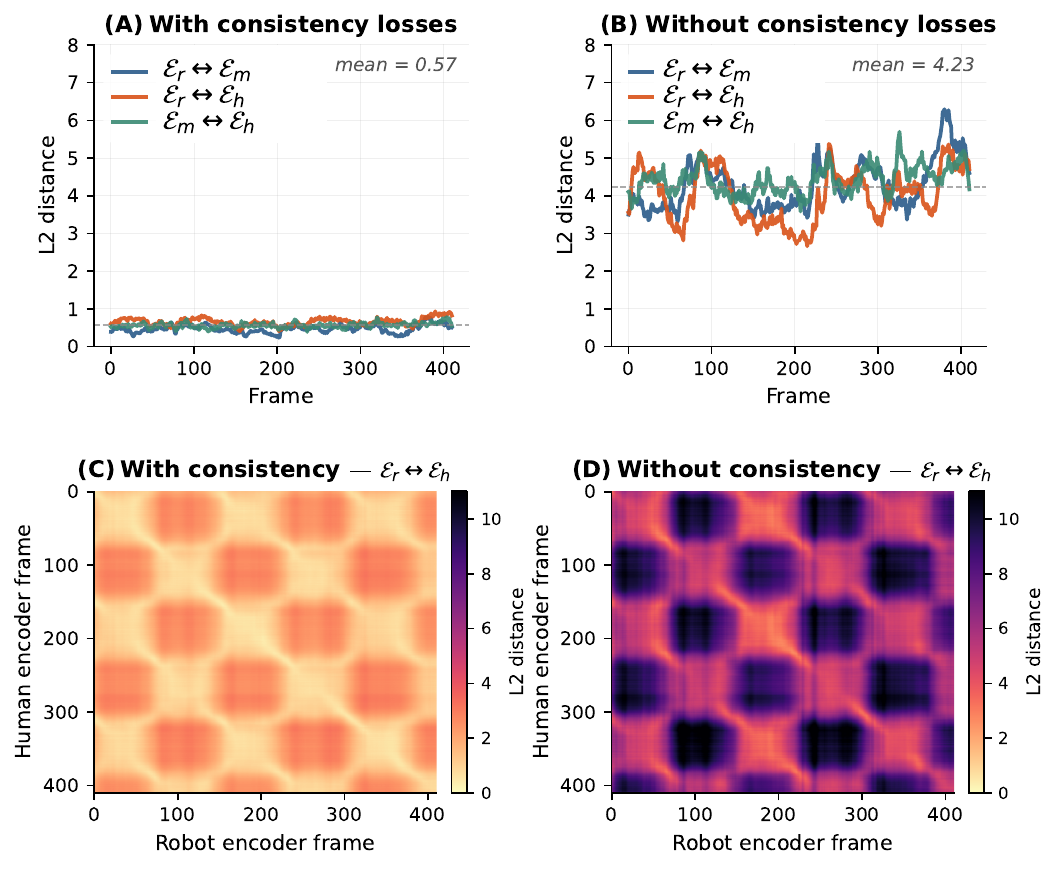}
    \caption{\textbf{Consistency losses align the multi-encoder latent space.} Latent space alignment with and without the consistency losses, visualized on a crawling motion. \textbf{(A and B)} Diagonal L2 distance between encoder pairs. \textbf{(C and D)} Cross-encoder distance matrices. With the consistency losses, matching frames produce aligned tokens; without them, the alignment breaks down.}
    \label{fig:latent_alignment}
\end{figure}

Our multi-encoder design mapped heterogeneous input types (robot motion, human SMPL poses, and hybrid teleop commands) into a shared token space, aligned by the consistency losses $\mathcal{L}_{\text{token}}$ and $\mathcal{L}_{\text{cycle}}$ (\secref{sec:ablations}). \Cref{fig:latent_alignment} visualizes this alignment on a crawling motion, with and without the consistency losses. Removing the consistency losses caused an $8\times$ increase in cross-encoder divergence, confirming that they were necessary for cross-encoder alignment.

\subsection{Supplementary Videos} \label{app:videos}

The \href{https://nvlabs.github.io/GEAR-SONIC/}{project site} hosts a collection of high-resolution videos that complement the results in the main paper, demonstrating large-scale motion tracking, interactive control, multi-modal interfaces, teleoperation, and foundation-model-driven loco-manipulation on a Unitree G1. The videos are provided as Supplementary Movies~S1 to S5 and are cited in the main text in order of appearance.

\subsubsection{\href{https://youtu.be/2BWl2uAO5lc}{Movie~S1: Summary Video}}
These clips demonstrate universal motion tracking across diverse behaviors, together with robustness under strong external perturbations: an operator drops an approximately 11\,kg (25\,lb) object onto the robot during policy execution, and the robot absorbs the impact, maintains balance, and continues tracking without any recovery module or policy adaptation (see \secref{app:robustness}).

\subsubsection{\href{https://nvlabs.github.io/GEAR-SONIC/\#interactive-kinematic-planner}{Movie~S2: Interactive Motion Control}}
These videos demonstrate the real-time kinematic motion planner (\secref{sec:method:planner}) coupled with the universal tracking policy.

\paragraph{Stylized Locomotion}
Users control the humanoid through continuous velocity and heading commands via gamepad. The planner replans at 10\,Hz, producing responsive and natural movements. Clips show smooth transitions between motion styles: happy walk, running, stealth walk, injured walk, and sprinting.

\paragraph{Interactive Boxing}
\shortname executes jabs, hooks, stance shifts, and reactive upper-body movements as continuous, unbroken sequences driven by the kinematic planner.

\paragraph{Squatting, Kneeling, and Crawling}
Videos show arbitrary pelvis-height control (0.3--0.8\,m), kneeling variations, elbow crawling, hand-knee crawling, and transitions. These skills enable locomotion in confined spaces and downstream teleoperation behaviors.

\subsubsection{\href{https://nvlabs.github.io/GEAR-SONIC/\#multi-modal-control}{Movie~S3: Video and Multi-Modal Control}}

\paragraph{Video Teleoperation}
Using video as input and GEM~\citep{genmo2025} for pose estimation at $\geq$60\,fps, the humanoid tracks and reproduces complex motions from human demonstrations in real time. Clips include walking, running, turning, jumping, dancing, and floor motions.

\paragraph{Text-Driven Motion}
\shortname responds to natural-language commands such as ``walk forward'', ``punch with your left hand'', or ``act like a monkey'' and can track the generated motion from GEM~\citep{genmo2025}.

\paragraph{Music-Conditioned Motion}
GEM~\citep{genmo2025} generates dance motions conditioned on melodic and rhythmic structure; \shortname tracks the generated choreography in real time. Both text and music modalities use the same universal control policy with no retraining.

\subsubsection{\href{https://nvlabs.github.io/GEAR-SONIC/\#teleoperation}{Movie~S4: VR Teleoperation}}

\paragraph{Full-Body VR Control}
Using the PICO whole-body motion-tracking interface~\citep{zhao2025xrobotoolkit} (headset, ankle trackers, and controllers), full-body SMPL pose streams drive \shortname via the human-motion encoder $\bs{\mathcal{E}}_h$. The robot performs walking, sidestepping, crouching, and loco-manipulation with low latency.

\paragraph{3-Point VR Teleoperation}
A lightweight teleoperation mode using only the headset and handheld controllers (no ankle trackers). The user provides upper-body SE(3) poses via the controllers; the kinematic planner generates the lower body. Videos show mobile manipulation, navigation, reaching, and object interaction tasks.

\subsubsection{\href{https://nvlabs.github.io/GEAR-SONIC/\#vla-foundation-model}{Movie~S5: Foundation-Model-Driven Loco-Manipulation}}
These clips show a GR00T N1.5 VLA model~\citep{gr00t-n1_5-blog, bjorck2025gr00t} driving \shortname via the universal token interface. Videos cover all five VLA tasks: apple to plate (3-point interface), object pickup (carrot, scrub), open trash can with foot, soda can to trash can (multi-step: pick up, navigate, step on pedal, throw), and drill and box relocation. All policies are fully autonomous with no human intervention during execution.

\subsection{Supplementary Data}

Tabulated source data underlying the quantitative main-text figure panels are provided as a separate machine-readable file, Data S1 (a multi-sheet spreadsheet with one sheet per figure panel).

\end{document}